\documentclass{article} % For LaTeX2e
\usepackage{iclr2025_conference,times}
\usepackage{amsmath,amsfonts,bm}

\def\eqref#1{equation~\ref{#1}}
\def\1{\bm{1}}

\DeclareMathAlphabet{\mathsfit}{\encodingdefault}{\sfdefault}{m}{sl}
\SetMathAlphabet{\mathsfit}{bold}{\encodingdefault}{\sfdefault}{bx}{n}

\usepackage{hyperref}
\usepackage{url}
\usepackage{inconsolata}

\usepackage{times}
\usepackage{latexsym}
\usepackage{amsfonts}
\usepackage{amsmath}
\usepackage{amssymb}
\usepackage{multicol}
\usepackage{multirow}
\usepackage{xspace}
\usepackage{booktabs}
\usepackage{bbding}
\usepackage{array}
\usepackage{threeparttable}
\usepackage{tcolorbox}
\usepackage{tabularx}
\usepackage{enumitem}
\usepackage{xcolor,colortbl}
\usepackage{color}
\usepackage{setspace}
\usepackage{makecell}
\usepackage{listings}

\usepackage{xltabular}
\usepackage{tcolorbox}
\tcbuselibrary{breakable}
\usepackage{xcolor}

\lstdefinestyle{python}{
    language=Python,
    basicstyle=\ttfamily\small,
    keywordstyle=\color{blue}\bfseries,
    commentstyle=\color{green},
    stringstyle=\color{red},
    numberstyle=\tiny\color{gray},
    showstringspaces=false,
    frame=single,
    breaklines=true,
    backgroundcolor=\color{lightgray!20}
}

\usepackage{booktabs} % 用于美化表格的横线

\usepackage[utf8]{inputenc} % allow utf-8 input
\usepackage[T1]{fontenc}    % use 8-bit T1 fonts
\usepackage{hyperref}       % hyperlinks
\usepackage{url}            % simple URL typesetting
\usepackage{booktabs}       % professional-quality tables
\usepackage{amsfonts}       % blackboard math symbols
\usepackage{nicefrac}       % compact symbols for 1/2, etc.
\usepackage{microtype}      % microtypography
\usepackage{xcolor}         % colors
\usepackage{float}
\usepackage{graphicx}       % for table and figure
\usepackage{wrapfig}        % 用于环绕文本图表
\usepackage{subcaption}     % 用于环绕文本表格的标题
\usepackage{makecell}
\usepackage{soul}

\hypersetup{
    colorlinks=true,
    linkcolor=blue,
    citecolor=blue,
}

\definecolor{Ocean}{RGB}{129,194,250}

\definecolor{deepgreen}{RGB}{0, 70, 0}
\usepackage{cleveref}
\crefformat{section}{\S#2#1#3}
\crefformat{subsection}{\S#2#1#3}
\crefformat{subsubsection}{\S#2#1#3}
\crefrangeformat{section}{\S\S#3#1#4 to~#5#2#6}
\crefmultiformat{section}{\S\S#2#1#3}{ and~#2#1#3}{, #2#1#3}{ and~#2#1#3}
\crefmultiformat{subsection}{\S\S#2#1#3}{ and~#2#1#3}{, #2#1#3}{ and~#2#1#3}
\Crefformat{figure}{#2Fig.~#1#3}
\Crefmultiformat{figure}{Figs.~#2#1#3}{ and~#2#1#3}{, #2#1#3}{ and~#2#1#3}
\Crefformat{table}{#2Tab.~#1#3}
\Crefmultiformat{table}{Tabs.~#2#1#3}{ and~#2#1#3}{, #2#1#3}{ and~#2#1#3}
\Crefformat{appendix}{Appx.~\S#2#1#3}
\crefmultiformat{appendix}{Appx.~\S#2#1#3}{ and~#2#1#3}{, #2#1#3}{ and~#2#1#3}
\crefformat{algorithm}{Alg.~#2#1#3}
\Crefformat{equation}{Eq.~#2#1#3}

\makeatother
\newcommand{\mytextbox}[2]{\tikzmarknode[draw=#1,thick,inner sep=2pt]{test}{\myfontsize #2}}
\definecolor{myred}{rgb}{0.7, 0.3, 0.0}
\definecolor{myblue}{HTML}{054488}
\definecolor{mygreen}{HTML}{056b34}
\definecolor{myorange}{HTML}{ff8800}
\definecolor{mypurple}{HTML}{8400ff}
\definecolor{mypink}{HTML}{f7acb9}

\newcommand{\red}[1]{\mytextbox{myred}{\textbf{\textcolor{myred}{#1}}}}
\newcommand{\blue}[1]{\mytextbox{myblue}{\textbf{\textcolor{myblue}{#1}}}}
\newcommand{\green}[1]{\mytextbox{mygreen}{\textbf{\textcolor{mygreen}{#1}}}}

\newcommand{\purple}[1]{\mytextbox{mypurple}{\textbf{\textcolor{mypurple}{#1}}}}

\definecolor{myred}{rgb}{0.7, 0.3, 0.0}
\definecolor{myblue}{HTML}{054488}
\definecolor{mygreen}{HTML}{056b34}

\usepackage{times}
\usepackage{CJKutf8}
\usepackage{latexsym}
\usepackage{tcolorbox}
\usepackage{multirow}
\usepackage{wrapfig}
\usepackage{tikz}
\usepackage{capt-of}
\usepackage{graphicx}  %Required
\usepackage{pgfplots}
\pgfplotsset{compat=1.12}
\usepackage{amsmath}
\usepackage{multicol}
\usepackage{color}
\usepackage{mwe}
\usepackage{wrapfig}
\usepackage{colortbl,array}
\usepackage{xspace}
\usepackage{tikz}
\usetikzlibrary{tikzmark}
\makeatletter
\newcommand*\myfontsize{%
  \@setfontsize\myfontsize{7}{8}%
}
\makeatother
\definecolor{myred}{rgb}{0.7, 0.3, 0.0}
\definecolor{myblue}{HTML}{054488}
\definecolor{mygreen}{HTML}{056b34}

\newcolumntype{R}[1]{>{\raggedleft\let\newline\\\arraybackslash\hspace{0pt}}m{#1}}

\usetikzlibrary{intersections}

\definecolor{darkgreen}{rgb}{0.0, 0.42, 0.24}
\usepackage{caption}
\usepackage{subcaption}
\usepackage{graphicx}
\usepackage{pifont}
\usepackage{titletoc}
\usepackage{amsfonts}
\usepackage{booktabs}
\usepackage{arydshln}
\usepackage{colortbl}
\usepackage{algorithm}
\usepackage[noend]{algpseudocode}
\usepackage{enumitem}
\usepackage{graphicx}
\usepackage{soul}
\usepackage{colortbl,array,xcolor}
\usepackage{amsmath}
\usepackage{fontawesome}

\title{Agentic Reinforced Policy Optimization}

\author{Guanting Dong$^1$\thanks{Work done during internship at Kuaishou, supervised by Hangyu Mao (\texttt{hy.mao@pku.edu.cn}).} \ , Hangyu Mao$^2$, Kai Ma$^2$, Licheng Bao$^2$\footnotemark[1] \ , Yifei Chen$^1$, Zhongyuan Wang$^2$\footnotemark[1]  \\
\textbf{Zhongxia Chen$^2$, Jiazhen Du$^2$, Huiyang Wang$^2$\footnotemark[1] \ , Fuzheng Zhang$^2$, Guorui Zhou$^2$$^\dagger$}   \\
\textbf{Yutao Zhu$^1$, Ji-Rong Wen$^1$, Zhicheng Dou$^1$\thanks{Corresponding author.}} \\
$^1$Renmin University of China, $^2$Kuaishou Technology\\
\texttt{\{dongguanting, dou\}@ruc.edu.cn}\\
\\
\begin{tabular}{@{}ll@{}}
\faGithub\ GitHub: \href{https://github.com/dongguanting/ARPO}{\texttt{\textcolor{cyan}{https://github.com/dongguanting/ARPO}}}
\end{tabular}
}

\iclrfinalcopy % Uncomment for camera-ready version, but NOT for submission.
\begin{document}

\maketitle
\vspace{-1em}
\begin{abstract}
\vspace{-0.5em}
Large-scale reinforcement learning with verifiable rewards (RLVR) has demonstrated its effectiveness in harnessing the potential of large language models~(LLMs) for single-turn reasoning tasks. In realistic reasoning scenarios, LLMs can often utilize external tools to assist in task-solving processes. However, current RL algorithms inadequately balance the models' intrinsic long-horizon reasoning capabilities and their proficiency in multi-turn tool interactions. To bridge this gap, we propose \textbf{A}gentic \textbf{R}einforced \textbf{P}olicy \textbf{O}ptimization~(\textbf{ARPO}), a novel \textit{agentic} RL algorithm tailored for training multi-turn LLM-based agents. Through preliminary experiments, we observe that LLMs tend to exhibit highly uncertain behavior, characterized by an increase in the entropy distribution of generated tokens, immediately following interactions with external tools. Motivated by this observation, ARPO incorporates an entropy-based adaptive rollout mechanism, dynamically balancing global trajectory sampling and step-level sampling, thereby promoting exploration at steps with high uncertainty after tool usage. By integrating an advantage attribution estimation, ARPO enables LLMs to internalize advantage differences in stepwise tool-use interactions. Our experiments across 13 challenging benchmarks in computational reasoning, knowledge reasoning, and deep search domains demonstrate ARPO's superiority over trajectory-level RL algorithms. Remarkably, ARPO achieves improved performance using only half of the tool-use budget required by existing methods, offering a scalable solution for aligning LLM-based agents with real-time dynamic environments.

\end{abstract}
\begin{figure}[H]
    \centering
    \vspace{-1em}
    \includegraphics[width=0.98\linewidth]{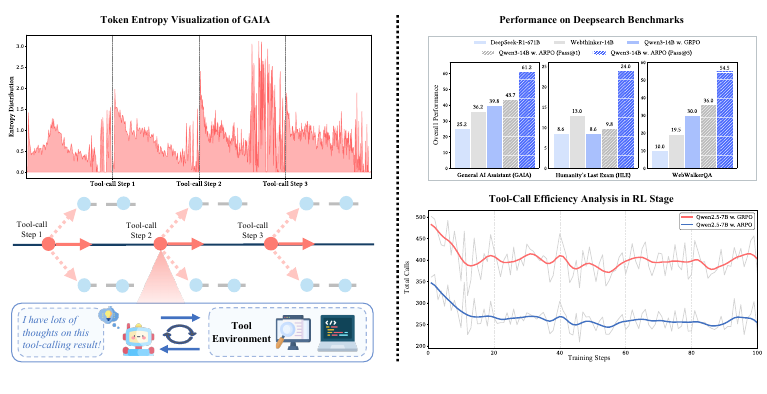}
    \vspace{-2em}
    \caption{Overview of tool-use token entropy exploration and ARPO algorithm performance. \textbf{Left:} High entropy observed in the LLM following tool usage. \textbf{Right:} LLM performance comparison on deep search tasks using \textit{\textbf{only 1k RL samples}}, along with a comparison of training tool-use budgets.}
    \label{fig:intro}
\end{figure}
\section{Introduction}

Recently, large-scale Reinforcement Learning with Verifiable Rewards (RLVR) has demonstrated strong potential in unleashing the capabilities of frontier large language models (LLMs), showcasing impressive performance across various  single-turn reasoning tasks~\citep{2409_openai_o1,deepseek-r1,team2025kimi,qwen2.5,qwen3,rft,skywork_1}. However, in open-ended reasoning scenarios~\citep{agentq,shridhar2020alfworld,toolllm,GAIA}, LLMs should not only cultivate long-horizon planning and adaptive decision-making skills, but also engage in dynamic, multi-turn interactions with external tool environments. To address these challenges, \textbf{Agentic} \textbf{R}einforcement \textbf{L}earning (\textbf{Agentic RL})~\citep{agentic_rl_1,mai2025agent}\footnote{\url{https://moonshotai.github.io/Kimi-K2/}} has emerged as a promising training paradigm,  shifting LLMs training from a static task-solving to the landscape of dynamic agent-environment reasoning~\citep{alphazero,dong2025toolstar,wen2024reinforcing,retool,qian2025toolrl,searchr1}.

Current agentic reinforcement learning (RL) methods typically employ trajectory-level algorithms like GRPO or DAPO~\citep{deepseekmath,yuDAPO,torl,li2025websailor,wu2025webdancer,r1searcher}. These approaches independently sample complete tool-use trajectories using predefined special tokens and provide reward signals based on the final output. To address tool overuse and sparse reward issues~\citep{smart}, several studies have attempted to design elegant reward functions to better align tool-use behavior~\citep{wang2025otc,sem,efficient_code_zhaoxin,wang2025acting}. Despite some progress, these optimizations often overlook a crucial aspect of training LLM-based agents: the multi-turn interaction loops between the LLM and the tool environment~\citep{ragen,mai2025agent,Group-in-Group}. Unlike the single-turn reasoning paradigm, multi-turn tool-use loops offer LLMs diverse and informative feedback in real-time. This characteristic underscores the necessity of discovering effective stepwise tool-use behaviors.

To gain an insight into the step-level tool-use behavior of LLMs, we draw inspiration from a series of entropy-based RL studies~\citep{20/80,wang2025reinforcement,zhaoxin_entropy,zheng2025first} and quantify the token entropy distribution of LLM-based search agents during deep search tasks. As illustrated in Figure~\ref{fig:intro} (left), the initial tokens generated by the LLM after receiving each round of tool-call feedback consistently exhibit a high entropy. This indicates that external tool-call significantly introduces uncertainty into the LLM's reasoning process, uncovering latent behaviors that remain under-explored by LLM-based agents~\citep{ruan2023tptu,kong2024tptu,li2025ngent,Group-in-Group,latent_reasoning_survey}. Unfortunately, current trajectory-level RL methods often overemphasize complete roll-out sampling comparisons, neglecting the exploration of fine-grained behavior at each tool-use step~\citep{step_agent,yu2024steptool,zhang2025r1}. This oversight limits the diversity and scope necessary for aligning better tool-use behaviors. Consequently, it is essential to develop an agentic RL algorithm that aligns with agent-environment interaction characteristics to fully realize the potential of LLM-based agents.

To bridge this gap, we propose \textbf{A}gentic \textbf{R}einforced \textbf{P}olicy \textbf{O}ptimization (\textbf{ARPO}), an agentic RL algorithm tailored for training multi-turn LLM-based agent. The core principle of ARPO is to encourage the policy model to adaptively branch sampling during high-entropy tool-call rounds, thereby efficiently aligning step-level tool-use behaviors. In detail, we propose an entropy-based adaptive rollout mechanism that integrates both global and partial sampling perspectives. In the rollout phase, the LLM initially performs multiple global samplings , recording the initial entropy distribution of each sample. After each tool-calling, we further monitor the real-time token entropy variation, regarding them as branching criteria.
If the entropy change exceeds a predefined threshold, the model performs additional partial sampling to explore more diverse tool-integrated reasoning behaviors. This design allows ARPO to effectively expand the original sampling space while balancing global and step-level tool-use behavior learning.

To fully exploit the benefits of adaptive sampling, we introduce the advantage attribution estimation. Specifically, we explores both hard and soft advantage settings of ARPO, assigning shared advantage values to tokens along the same source reasoning path, while tokens on branched paths receive distinct advantage values. This mechanism encourages the model to internalize advantage differences in stepwise tool-use behaviors. 

% Beyond heuristic motivation, we also theoretically demonstrate the rationale of applying the ARPO algorithm in LLM-based agent training.

In our experiments, we comprehensively evaluate 13 datasets across computational reasoning, knowledge reasoning, and deep search domains. Figure~\ref{fig:intro} (right) provides an overview of the results for the deep search tasks. ARPO consistently surpasses traditional sample-level RL algorithms in agentic training. Remarkably, ARPO achieves this with only half the tool-call budget required by trajectory-level RL methods, striking an optimal balance between accuracy and efficiency. Further scaling analyses validate ARPO's capacity to enhance LLM's agentic reasoning in a scalable manner.

In summary, the key contributions of this work are as follows:
\begin{itemize}[leftmargin=1em]
\item We quantify the token entropy variation of LLM during agentic reasoning, revealing the inherent limitations of trajectory-level RL algorithms for aligning LLM-based agents.

\item We propose the ARPO algorithm, which integrates an entropy-based adaptive rollout mechanism to balances global sampling while encouraging branch sampling during high-entropy tool-use steps. Furthermore, ARPO employs Advantage Attribution Estimation to assist the LLM in better internalizing advantage differences in stepwise tool-use behaviors.

\item Beyond heuristic motivation, we also theoretically demonstrate the rationale of applying the ARPO algorithm in LLM-based agent training.

\item Experiments across 13 challenging benchmarks consistently demonstrate ARPO's superiority over mainstream RL algorithms, requiring only half the tool-use training budgets, thereby offering practical insights for exploring agentic RL algorithms.
\end{itemize}

% \textbf{Tool-Integrated Reasoning Agent.}

\section{Preliminary}

Before introducing ARPO, we first provide a brief overview of key concepts and review preliminary entropy-based experiments on LLM reasoning with tools.
\subsection{Agentic Reinforcement Learning}

In this section, we formulate the agentic RL training objective as:
\begin{equation}
\max_{\pi_\theta} \mathbb{E}_{x \sim \mathcal{D}, y \sim \pi_\theta(\cdot \mid x; T)} 
\left[ r_\phi(x, y) \right] 
- \beta \, \mathbb{D}_{\text{KL}} \left[ \pi_\theta(y \mid x; T) \,\|\, \pi_{\text{ref}}(y \mid x; T) \right],
\end{equation}
where $T$ denotes the set of available tools, $\pi_{\theta}$ represents the policy LLM, $\pi_{\text{ref}}$ is the reference LLM, $r_{\phi}$ and $\mathbb{D}_{\text{KL}}$ denotes the reward function and KL divergence respectively. The input $x$ is sampled from dataset $\mathcal{D}$, and $y$ is the corresponding output, possibly interleaved with tool-call feedback.

Unlike conventional RL methods that rely solely on LLM rollouts, agentic RL incorporates tool-call feedback during the reasoning process~\citep{chat_cot,tora,torl,wu2025agentic,dotamath}. The rollout sampling can be decomposed as:
\begin{equation}
P_\theta(\mathcal{R}, y \mid x; T) = 
\underbrace{\prod_{t=1}^{t_{\mathcal{R}}} P_\theta(\mathcal{R}_t \mid \mathcal{R}_{<t}, x; T)}_{\text{Agentic Reasoning}} 
\cdot 
\underbrace{\prod_{t=1}^{t_y} P_\theta(y_t \mid y_{<t}, \mathcal{R}, x; T)}_{\text{Answer Generation}},
\label{eq:tir}
\end{equation}
where $\mathcal{R}$ is the reasoning trajectory of length $t_{\mathcal{R}}$, interleaved with tool-call feedback, and $y$ is the final answer with length $t_y$. Our ARPO is built upon rule-based RL algorithm (e.g. GRPO~\citep{deepseekmath}, Reinforce++~\citep{hu2025reinforce++}) designed to optimize LLM-based agents.

\begin{figure*}[t]
    \centering
    \small    \includegraphics[width=0.98\linewidth]{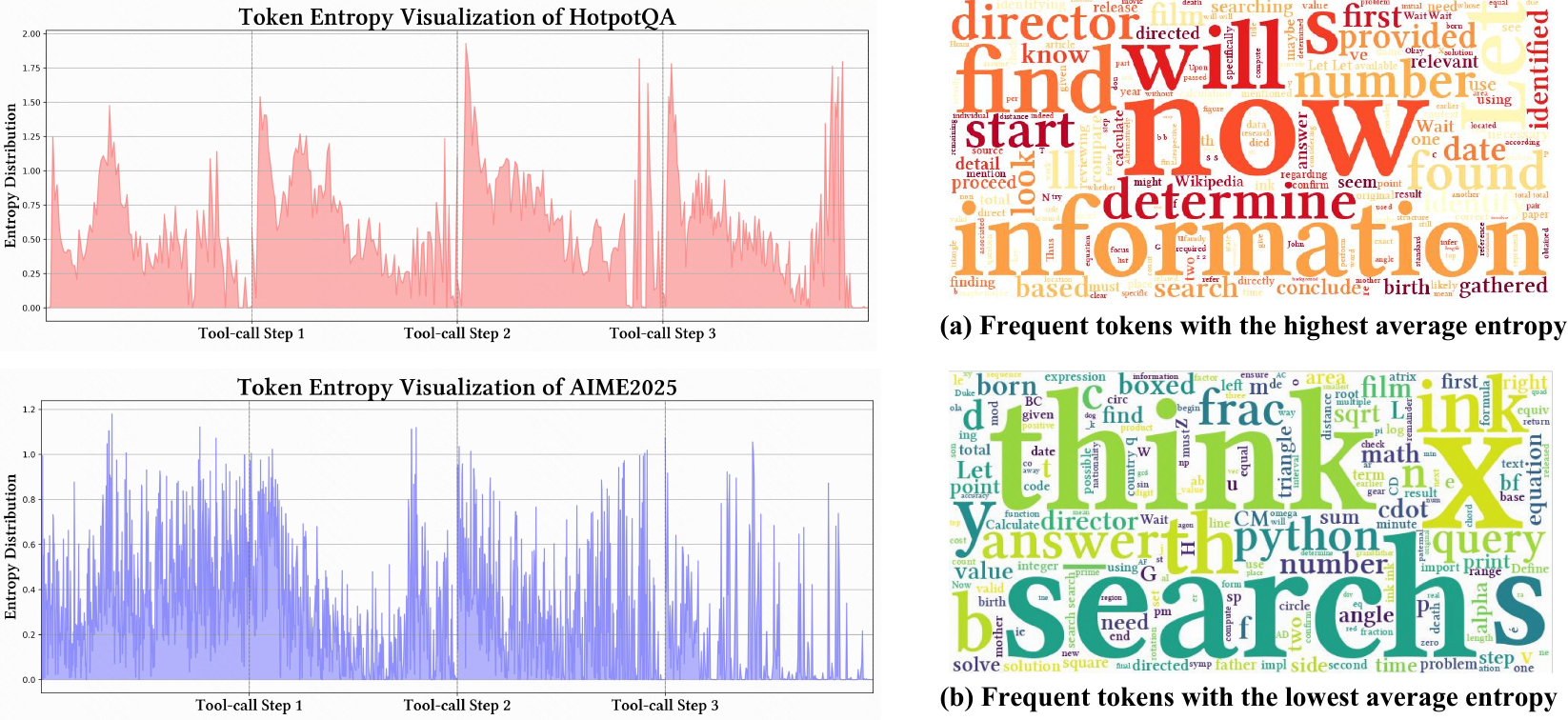}
    \vspace{-0.5em}
    \caption{Analysis of token entropy variations and token frequency statistics of LLM-based tool-use agent across different datasets.}
    \label{main1}
    % \vspace{-1.5em}
\end{figure*}

\subsection{Analyzing Token Entropy in Agentic Reasoning}
\label{sec:pre-experiment}
\textbf{Token Entropy Calculation.} Following recent entropy-based RL studies~\citep{20/80,wang2025reinforcement,zhaoxin_entropy,zheng2025first}, we compute the token-level generation entropy at step $t$ as:
\begin{equation}
H_t = -\sum_{j=1}^{V} p_{t, j} \log p_{t, j}, \quad \text{where } \boldsymbol{p}_t = \pi_\theta\left(\cdot \mid \mathcal{R}_{<t}, x; T\right) = \operatorname{Softmax}\left(\frac{\boldsymbol{z_t}}{\tau}\right).
\label{eq:entropy_cacluate}
\end{equation}

Here, $V$ is the vocabulary size, $\boldsymbol{z_t} \in \mathbb{R}^V$ is the pre-softmax logits, and $\tau$ is the decoding temperature. Note that this entropy reflects the uncertainty in the token generation distribution, rather than the uncertainty of any particular token.

\textbf{Pilot Experiment on Token Entropy.} To gain deeper insights into the reasoning process of LLM-based tool-use agents, we conduct a pilot study with two types of agents: one using a search engine for knowledge-intensive tasks and another using a Python interpreter for computational tasks. We measure token entropy variations throughout the reasoning process to assess uncertainty.

As shown in Figure \ref{main1}, our key observations are:
\textbf{(1)} Entropy rises sharply in the first 10–50 tokens following each tool call;
\textbf{(2)} Entropy tends to increase during early reasoning stages, but remains lower than after receiving tool-call feedback;
\textbf{(3)} Search feedback introduces more uncertainty than Python feedback.

We attribute these effects to the distributional shift between external feedback and the model’s internal reasoning (\textbf{Ob.1}), which introduces uncertainty often exceeding that of the original input (\textbf{Ob.2}). Furthermore, search engines typically return informative textual content, whereas Python outputs consist of deterministic numbers, resulting in greater entropy fluctuations in the former case (\textbf{Ob.3}).

These findings highlight a limitation of trajectory-level RL methods, which focus on initial reasoning while overlooking the uncertainty introduced by tool-call feedback. Our proposed ARPO algorithm addresses this by incorporating entropy-based exploration tailored to LLM agent training.

\subsection{Agentic Tool Design} 

In this work, we mainly focus on optimizing the training algorithms of LLM-based tool-use agents. After a comprehensive review of agentic RL studies~\citep{dong2025toolstar,retool,searchr1}, we identify three representative tools to empirically evaluate the effectiveness of ARPO:
\begin{itemize}[leftmargin=1em]
\item \textbf{Search Engine:} Retrieves relevant information by executing queries across the web.
\item \textbf{Web Browser Agent:} Accesses and parses relevant web links returned by the search engine, extracting and summarizing key content.
\item \textbf{Code Interpreter:} Automatically executes code generated by the language model, returning execution results if successful, or compiler error messages otherwise.
\end{itemize}

\section{Agentic Reinforce Policy Optimization}

\begin{figure}[t]
    \centering
    \includegraphics[width=1\linewidth]{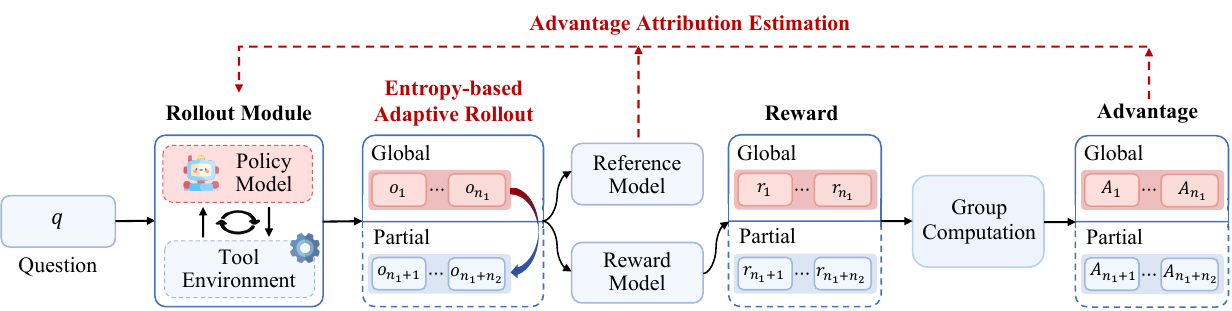}
    \vspace{-1em}
    \caption{The overview of ARPO algorithm.}
    \vspace{-1em}
    \label{arpo_1}
\end{figure}

In this section, we propose the ARPO algorithm, designed to guide LLMs in exploring step-wise tool-use behaviors under entropy-based guidance, as illustrated in Figures~\ref{arpo_1} and ~\ref{arpo_2}:
\begin{itemize}[leftmargin=1em]
\item \textbf{Entropy-based Adaptive Rollout} (\cref{sec:partial_rollout}): Inspired by the entropy variations observed in preliminary experiments (\cref{sec:pre-experiment}), ARPO extends the traditional rollout process by performing not only trajectory-level sampling but also branching at high-entropy tool-use steps. By striking a balance between global and partial sampling, ARPO encourages broader exploration of tool-use behaviors.

\item \textbf{Advantage Attribution Estimation} (\cref{sec:advantage_estimation}): To better accommodate the adaptive rollout mechanism, we propose the advantage attribution estimation, enabling the model to more effectively internalize the advantage differences in stepwise tool-use behaviors.

\item \textbf{Theoretical Analysis} (\cref{sec:Theoretical Foundation}): To establish the theoretical foundation of ARPO, we provide a formal analysis showing that ARPO offers good adaptability in multi-turn training scenario for LLM-based agents.
\end{itemize}
Below, we will delve into the specifics of our approach.

\subsection{Entropy-based Adaptive Rollout.} 
\label{sec:partial_rollout}
Inspired by preliminary experiments (\Cref{sec:pre-experiment}), we incorporate both trajectory-level sampling and entropy-based partial sampling during the rollout phase to cover a more comprehensive sampling scope. The design of this mechanism involves the following four core steps:

\textbf{(1) Rollout Initialization:} Given a global rollout size of $M$, the LLM first generates $N$ trajectories via trajectory-level sampling based on the input question $q$, while the remaining $M - N$ trajectories budgets are reserved for partial sampling. We then compute the entropy of the first tokens $k$ in each trajectory using Equation~\ref{eq:entropy_cacluate}, forming the initial entropy matrix denoted as $H_{\text{initial}} \in \mathbb{R}^{1 \times k}$.

\textbf{(2) Entropy Variation Monitoring:} After recording the initial entropy, the model perform agentic reasoning with tools, as defined in Equation~\ref{eq:tir}. To continuously monitor the entropy dynamics following each tool invocation, we allow the model to generate $k$ additional tokens after concatenating the response from the tool call. For the tool-call step $t$, we compute a step-level entropy matrix denoted as $H_{t} \in \mathbb{R}^{1 \times k}$. We then quantify the normalized change in entropy relative to the initial state using the following formulation:
\begin{equation}
\Delta H_{t} = \text{Normalize}(H_{t} - H_{\text{initial}}) %, \quad \text{where} \ \{H_{t}, H_{\text{initial}} \}\in \mathbb{R}^{1 \times k}.
\end{equation}
where normalization means summing all the values of $\Delta H$ in dividing by the vocab size $V$. A positive $\Delta H$ indicates an increase in uncertainty after tool-call step $k$, whereas a negative value reflects a reduction in uncertainty.

\begin{figure}[t]
    \centering
    \includegraphics[width=\linewidth]{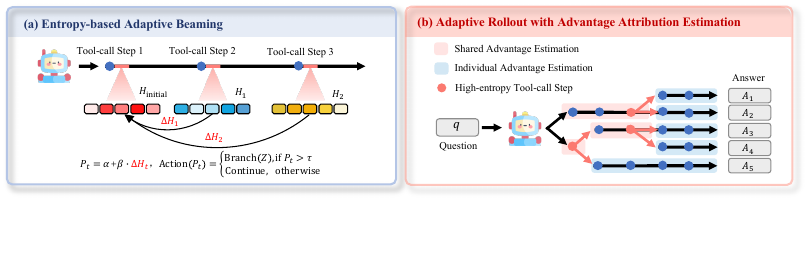}
    \vspace{-5.5em}
    \caption{llustration of two core components: Entropy-Based Adaptive Rollout and Advantage Attribution Estimation. \textbf{Left:}  Principle of Entropy-Based Adaptive Beaming. \textbf{Right:} ARPO assigns different advantages to shared and individual token parts in inter-group samples.}
    \label{arpo_2}
\end{figure}

\textbf{(3) Entropy-based Adaptive Beaming:} To encourage adaptive exploration along tool-use paths that exhibit beneficial entropy variations, we define the partial sampling probability at the tool-call step $t$ as follow:
\begin{equation}
P_t = \alpha + \beta \cdot \Delta H_t, \quad
\text{Action}(P_t) =
\begin{cases}
\text{Branch}(Z), & \text{if } P_t > \tau \\
\text{Continue}, & \text{otherwise}
\end{cases}
\end{equation}
where $\alpha$ is a base sampling probability, $\beta$ is a stability entropy value. As shown in Figure~\ref{arpo_2}(a), the model uses $P_t$ to determine its branching behavior: when $P_t$ exceeds a predefined threshold $\tau$, it initiates $\text{Branch}(Z)$, branching $Z$ partial reasoning paths from the current node; otherwise, it continues along the current trajectory. 

This mechanism enables the model to adaptively allocate exploration resources to regions of the reasoning space where rising entropy indicates a higher potential for informative outcomes.

\textbf{(4) Termination:} The process iterates until one of the conditions is satisfied: (1) if the total number of forked paths $\hat{Z}$ reaches the partial sampling budget $M - N$, branching stops and sampling continues until a final answer is produced; (2) if all paths terminate before reaching $M - N$, we supplement with $M - N - \hat{Z}$ additional trajectory-level samples to satisfy condition (1).

By leveraging this efficient rollout mechanism, ARPO facilitates uncertainty-aware exploration, allowing LLMs to more effectively identify step-level tool-calling behavior. Meanwhile, assuming the global expansion size and the number of tokens per trajectory are $n$, ARPO reduces the computational complexity of each rollout from the trajectory-level RL's $O(n^2)$ to between $O(n \log n)$ and $O(n^2)$ \footnote{Neglecting the minor overhead from token-level entropy calculations}.

\subsection{Advantage Attribution Estimation}
\label{sec:advantage_estimation}

Our entropy-based adaptive rollout mechanism naturally produces trajectories containing both shared reasoning token segments and distinct beam paths (Figure~\ref{arpo_2}), which motivates us to explore a more principled agentic RL policy update strategy. To this end, we consider the following two advantage assignment settings: 

\textbf{Hard Advantage Estimation:}  As shown in Figure~\ref{arpo_2}(b), a straightforward approach is to explicitly distinguish the shared and individual parts of each trajectory at the advantage level, thereby encouraging the model to capture step-level tool-use behaviors. Given $d$ trajectories that share certain tokens while diverging in others, we compute the advantage for the individual tokens using the normalized reward $R_i$: $\hat{A}_{i,t} = \frac{r_i - \text{mean}(\{R_i\}_{i=1}^G)}{\text{std}(\{R_i\}_{i=1}^G)}.$ For the shared tokens, we assign the average advantage across $d$ trajectories that contain the shared segment:
$\hat{A}_{i, t}^{\text{shared}} = \frac{1}{d} \sum_{i=1}^d \hat{A}_{i,t}.$

\textbf{Soft Advantage Estimation:} An elegant alternative to hard advantage assignment is to integrate the distinction between shared and individual token segments latently during policy optimization.  Specifically, for each input question $x$, the Group Relative Policy Optimization (GRPO)~\citep{deepseekmath} enables the reference policy $\pi_{\text{ref}}$ to generate a set of responses $\{y_1, y_2, \dots, y_G\}$ and optimizes the policy by maximizing:
\begin{small}
\begin{equation}
\begin{split}
J_{\text{GRPO}}(\theta) = \mathbb{E}_{(q,a) \sim D, \{y_i\}_{i=1}^G \sim \pi_{\theta_{\text{old}}}(\cdot \mid q)} \Bigg[ \frac{1}{G} \sum_{i=1}^G \frac{1}{|y_i|} \sum_{t=1}^{|y_i|} \min \Big( r_{i,t}(\theta) \hat{A}_{i,t}, \\
& \hspace*{-10.5em} \text{clip} \left( r_{i,t}(\theta), 1 - \epsilon, 1 + \epsilon \right) \hat{A}_{i,t} \Big) - \beta D_{\text{KL}}(\pi_{\theta} \parallel \pi_{\text{ref}}) \Bigg].
\end{split}
\label{eq:grpo}
\end{equation}
\end{small}
Notably, the GRPO objective incorporates the distinction between shared and individual tokens through importance sampling ratio $r_{i,t}(\theta)$:
% \begin{equation}
% r_{i,t}(\theta)= \frac{\pi_\theta(y_{i,t} \mid x, y_{i,<t})}{\pi_{\text{ref}}(y_{i,t} \mid x, y_{i,<t})}. 
% \end{equation}
\begin{equation}
\label{eq:r_ratio_importance}
r_{i,t}(\theta)= \frac{\pi_\theta(y_{i,t} \mid x, y_{i,<t})}{\pi_{\text{ref}}(y_{i,t} \mid x, y_{i,<t})}, \ 
\left\{
\begin{aligned}
& r_{i,t}(\theta) = r_{j,t}(\theta),\ && \text{if } y_{i,<t} = y_{j,<t} \ \text{(i.e., shared tokens)} \\
& r_{i,t}(\theta) \neq r_{j,t}(\theta),\ && \text{if } y_{i,<t} \neq y_{j,<t} \ \text{(i.e., individual tokens)}
\end{aligned}
\right.
\end{equation}
As indicated by the above equation, when trajectories $y_i$ and $y_j$ undergo a partial rollout at token $t$, they share the same response prefix tokens, i.e., $y_{i,<t} = y_{j,<t}$. Consequently, the shared prefix tokens in both trajectories are assigned the same importance weight $r_{i,t}(\theta)$. In the GRPO formulation, the mathematical interpretation is that the policy update is guided by the average advantage of tokens within each group, which serves as the loss signal. 
\begin{wrapfigure}[12]{r}{0.46\linewidth}
    \centering
    \includegraphics[width=0.99\linewidth]{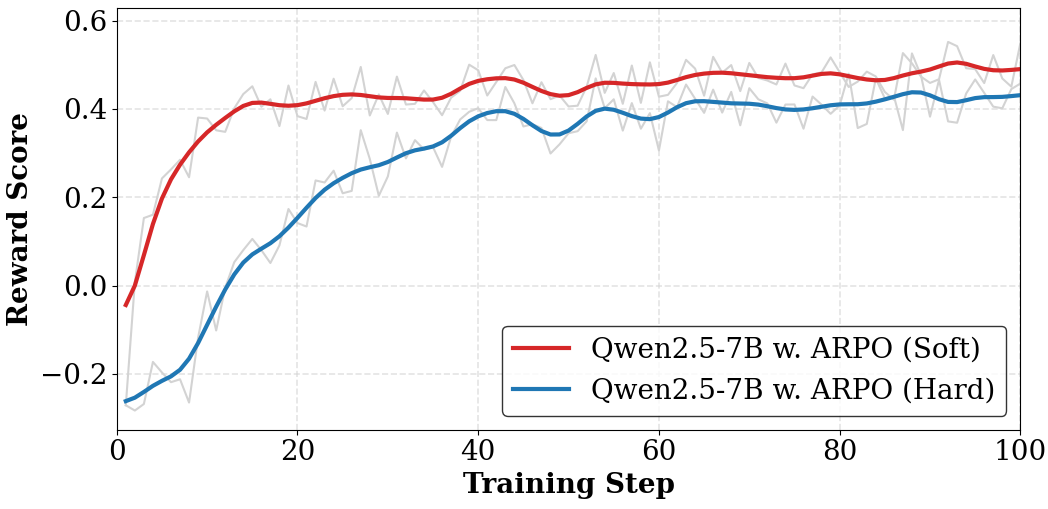}
    \vspace{-2em}
     \caption{Comparison of different advantage estimation method: Hard vs. Soft setting.}
    \label{fig:compare_adv}
\end{wrapfigure}
Since shared tokens have identical $r_{i,t}(\theta)$,, their advantage contributions are effectively aligned and closely approximate the advantage $\hat{A}_{i, t}^{\text{shared}}$ in a hard estimation setting. Although we adopt the GRPO loss formulation, our unique partial rollout design explicitly differentiates the update strategies for shared versus individual tokens. we also provide a detailed proof for the above argument in Appendix~\ref{app:Theoretical_proof}.

In practice, we further compare the reward variations between hard and soft advantage estimation in RL training. As shown in Figure~\ref{fig:compare_adv}, the soft setting achieves consistently higher rewards with greater stability during ARPO training. Consequently, our ARPO defaults to using the soft setting for advantage estimation.

\textbf{Hierarchical Reward Design.} The reward function serves as the optimization objective, guiding the policy model's behavior during training. We follow Tool-Star~\citep{dong2025toolstar}, considering both correctness and format rewards, along with a multi-tool collaboration reward mechanism. Notably, an additional reward $r_M$ is given when the model generates the correct answer, follows the correct tool invocation format, and uses multiple tools (i.e., \green{<search>} and \blue{<python>}) during reasoning. The overall reward $R$ is formally defined as:
\begin{equation}
\small
\hspace*{-1em} R = \begin{cases}
\max(Acc.+ r_{\text{M}}, Acc.) &\text{If Format is Good \& Acc.>0}  \\
0 & \text{If Format is Good \& Acc.=0 } \\
-1 & \text{Otherwise} \\
\end{cases}
,  r_{\text{M}} = 
\begin{cases}
0.1 & \text{If } \exists(\text{\green{<search>}} \&\text{\blue{<python>}}) \\
0 & \text{Otherwise}
\end{cases}
\end{equation}

The detailed flowchart for the ARPO algorithm can be found in Algorithm \ref{alg:arpo}.

\subsection{Theoretical Foundation}
\label{sec:Theoretical Foundation}
Our approach leverages the adaptive partial rollout mechanism, which involves branching at high-entropy tool-use steps. Here, we elucidate the rationale behind this mechanism.

As depicted in Figure \ref{arpo_2}, the adaptive partial rollout mechanism dynamically segments the Transformer-based policy's output tokens $< OT_1, OT_2, ..., OT_{|output|} >$ into $K$ segments. Each segment is defined as a macro action, $MA_i \triangleq < OT_m, OT_{m+1}, ..., OT_{m+n} >$. The corresponding macro states are defined as $MS_1 \triangleq < IT_1, IT_2, ..., IT_{|input|} >$ and $MS_i \triangleq < MS_{i-1}, MA_{i-1} >$. This segmentation allows us to derive the Generalized Policy Gradient (GPG) Theorem applicable to all Transformer-based policies:
\vspace{-0.5em}
\begin{equation}
\begin{aligned}
    \nabla_{\theta} J(\theta) 
    = & \mathbb{E}_{\tau \sim \pi_\theta}\{ \sum_{T=1}^{K} [\nabla_{\theta}\log\pi_{\theta}(MA_T|MS_T) A_T(\tau) ] \}
\end{aligned}
\end{equation}
\vspace{-0.5em}
In this equation, $T$ represents the macro step, and $A_T(\tau)$ denotes the advantage of trajectory $\tau$.

The GPG Theorem asserts that for any differentiable Transformer-based policy $\pi_{\theta}$ and any objective function $J(\theta)$, optimization can be effectively conducted using macro actions (i.e., partial rollout segments). This generalization encompasses the traditional Policy Gradient Theorem \citep{sutton1999policy}, $\nabla_{\theta} J(\theta) = \mathbb{E}_{\tau \sim \pi_\theta}\{ \sum_{t=1}^{H}[\nabla_{\theta}\log\pi_{\theta}(a_t|s_t) A_t(\tau)] \}$, which operates on single-token actions (where $a_t$ is a single output token of the Transformer), as a specific instance of our broader GPG framework. Consequently, ARPO as an advanced implementation of the GPG Theorem provides a robust theoretical foundation. Tthe formal proof of the GPG Theorem is presented in Appendix \ref{app:Theoretical_proof_GPG}.

\section{Experiment}

\subsection{Datasets.} 
To comprehensively evaluate the effectiveness of our ARPO algorithm in training LLM-based tool-using agents, we conduct experiments on the following three types of long-horizon reasoning tasks:

\textbf{(1) Mathematical Reasoning:} including AIME2024, AIME2025\footnote{\url{https://huggingface.co/datasets/AI-MO/aimo-validation-aime}}, MATH500 \citep{math500}, MATH \citep{MATH}, and GSM8K.

\textbf{(2) Knowledge-Intensive Reasoning:} including WebWalker \citep{2501_WebWalker}; as well as three Wikipedia-based open-domain QA tasks: HotpotQA \citep{hotpotqa}, 2WikiMultihopQA \citep{2wiki}, and Musique \citep{musique} and bamboogle~\citep{bamboogle}.

\textbf{(3) Deep Search:} including General AI Assistant (GAIA)~\citep{GAIA}, WebWalker \citep{2501_WebWalker}, Humanity’s Last Exam (HLE)~\citep{HLE}, and xbench~\citep{chen2025xbench}. Notably, we follow the Websailor~\citep{li2025websailor} setting by testing the xbench-DeepSearch split.

To maintain consistency with existing work, we use the Tool-Star~\citep{dong2025toolstar} test set split for the mathematical and knowledge reasoning benchmarks, and for the deep search benchmarks, we follow Webthinker and HIRA~\citep{webthinker,hira} for the Deepsearch test set split.

\subsection{Baselines.} To effectively evaluate the efficacy of ARPO, we consider the following three baselines:

\textbf{1. Direct Reasoning:} For mathematical and knowledge reasoning benchmarks, we evaluate the instruct versions of the Qwen2.5~\citep{qwen2.5} and Llama3.1~\citep{llama3} series. Given the superior mathematical performance of the Qwen3 series~\citep{qwen3}, we use the Deepsearch task to test RL algorithms on this model's backbone. We also reference strong reasoning models, including QwQ~\citep{qwen_qwq}, DeepSeek-R1~\citep{deepseek-r1}, GPT-4o~\citep{gpt_4o_system_card}, and o1-preview~\citep{gpt_4o_system_card}.

\textbf{2. Trajectory-level RL Algorithms:} We compare ARPO with common trajectory-level RL algorithms for training LLM-based tool-use agents, including GRPO~\citep{deepseekmath}, DAPO~\citep{yuDAPO}, and REINFORCE++~\citep{hu2025reinforce++}.

\textbf{3. LLM-based Search Agent:} For the deep search benchmark, we include GRPO and a series of open-source workflow-based search agents as references, such as vanilla RAG~\citep{rag_lewis}, Search o1~\citep{searcho1}, Webthinker~\citep{webthinker}, and ReAct~\citep{yao2022react}. Detailed introductions are in Appendix~\ref{app:baselines}.

\subsection{Training Guideline}
\label{sec:training_guide}

Our study aims to \textbf{validate the effectiveness of ARPO at the algorithmic level compared to traditional RL in training LLM agents, rather than merely pursuing performance improvements.} To ensure reproducibility, all training frameworks and datasets are sourced from open-access resources. Specifically, our experiments adhere to the cold-start SFT with RL paradigm~\citep{r1searcher,dong2025toolstar} to mitigate reward collapse during the initial RL training phases.

\textbf{1. Cold-Start Finetuning Phase:} Utilizing the LLaMAFactory~\citep{zheng2024llamafactory} framework, we leverage Tool-Star's open-source dataset of 54K training samples. To enrich the quality of mathematical reasoning data, we incorporate the STILL dataset (0.8K), drawing inspiration from CORT~\citep{cort}.

\textbf{2. RL Phase:} To assess ARPO across various scenarios, we explore the following domains:
\begin{itemize}[leftmargin=1em]
\item \textbf{Deep Reasoning Tasks:} This includes computational reasoning (e.g., AIME24, MATH500) and multi-hop knowledge-based reasoning (e.g., HotpotQA, Bamboogle). We utilize Tool-Star's 10K open-source RL training samples for algorithmic comparison.
\item \textbf{Deep Search Tasks:} These tasks require extensive web exploration and information integration, necessitating longer contexts and frequent tool interactions. We use \textbf{only 1K mixed hard search samples} from SimpleDeepSearcher~\citep{SimpleDeepSearcher} and WebSailor~\citep{li2025websailor} for training.
\end{itemize}
To expedite the RL training phase, we incorporate top-10 snippets from the Bing search engine as search results, employ a Python compiler within a sandbox environment, and use token-level F1 scores as the correctness signal.

\subsection{Evaluation Metric} 
In evaluation phrase, we use a search engine with browser capabilities to align with standard reasoning performance. For accuracy, we use F1 scores as the metric for four QA tasks in Knowledge-Intensive Reasoning, while other tasks are evaluated using Qwen2.5-72B-instruct under the LLM-as-Judge setting. We adopt pass@1 evaluation with non-zero temperature, setting the temperature and top-p to 0.6 and 0.95, respectively.For all tasks, we follow previous work~\citep{searcho1} and extract answers from the model output enclosed in \texttt{\textbackslash box\{\}}.

\begin{table*}[t]
\centering
\caption{Overall performance on 10 challenging reasoning tasks are presented. The top two outcomes are \textbf{bolded} and \underline{underlined}. Dataset abbreviations are as follows: HQA (HotpotQA), 2Wiki. (2wikiMultiHopQA), MuSi. (MuSiQue), and Bamb (Bamboogle).}
\vspace{-0.5em}
\label{tab:main_table}
\setlength\tabcolsep{1.5pt}
\renewcommand{\arraystretch}{1}
\fontsize{8.1pt}{10.5pt}\selectfont
\begin{tabular}{p{2.8cm}ccccccccccc}
\toprule
\multirow{2}[2]{*}{\textbf{Method}} & \multicolumn{5}{c}{\textbf{Mathematical Reasoning}} & \multicolumn{5}{c}{\textbf{Knowledge-Intensive Reasoning}} & \multirow{2}[2]{*}{\textbf{Avg.}} \\
\cmidrule(lr){2-6} \cmidrule(lr){7-11}
& AIME24 & AIME25 & MATH500 & GSM8K & MATH & WebWalker & HQA & 2Wiki. & MuSiQ. & Bamb. & \\
\midrule
\textbf{Qwen2.5-3B-Instruct} & 10.0 & 6.7 & 63.0 & 75.0 & 71.6 & 0.5 & 9.7 & 9.4 & 3.6 & 11.7 & 26.1 \\
\quad + TIR Prompting                 & 6.7 & 6.7 & 52.2 & 56.6 & 62.8 & 14.0 & 15.4 & 14.1 & 6.1 & 16.4 & 25.1 \\
\quad + GRPO                 & \underline{20.0} & 13.3 & \textbf{72.0} & \textbf{86.0} & 81.0 & \underline{21.0} & \underline{56.5} & \underline{64.5} & 24.7 & 65.2 & 50.4 \\
\quad + Reinforce ++         & 16.7 & 13.3 & 70.4 & \underline{85.0} & 80.2 & 19.5 & 55.9 & 62.3 & 27.9 & \underline{65.7} & 49.7 \\
\quad + DAPO                 & \underline{20.0} & \underline{16.7} & 71.2 & \underline{85.0} & \underline{81.2} & 19.5 & 54.8 & 62.5 & \textbf{30.0} & 64.8 & \underline{50.6} \\
\rowcolor[RGB]{236,244,252} 
\quad + ARPO                 & \textbf{23.3} & \textbf{20.0} & \underline{71.4} & \underline{85.0} & \textbf{82.5} & \textbf{24.5} & \textbf{58.5} & \textbf{67.4} & \underline{28.7} & \textbf{66.8} & \textbf{52.8} \\

\midrule
\textbf{Llama3.1-8B-Instruct} & 3.3 & 0.0 & 43.3 & 81.4 & 60.6 & 3.0 & 24.3 & 24.6 & 10.4 & 40.0 & 28.8 \\
\quad + TIR Prompting        & 3.3 & 3.3 & 39.4 & 73.8 & 58.2 & 15.0 & 48.5 & 47.5 & 15.5 & 58.4 & 36.3 \\
\quad + GRPO                 & 13.3 & \underline{13.3} & \underline{62.4} & \underline{87.4} & \underline{79.2} & 26.5 & \underline{57.8} & \underline{71.8} & \underline{31.0} & 68.2 & \underline{51.1} \\
\quad + Reinforce ++         & 13.3 & \textbf{16.7} & 61.4 & 87.0 & 77.2 & \underline{27.5} & 57.1 & 71.6 & 29.9 & \underline{69.1} & \underline{51.1} \\
\quad + DAPO                 & \underline{16.7} & \underline{13.3} & 61.2 & \underline{87.4} & 76.4 & 25.5 & 56.6 & 70.3 & 29.2 & 67.3 & 50.4 \\
\rowcolor[RGB]{236,244,252} 
\quad + ARPO                 & \textbf{23.3} & \textbf{16.7} & \textbf{64.6} & \textbf{88.0} & \textbf{80.2} & \textbf{30.5} & \textbf{65.4} & \textbf{75.5} & \textbf{34.8} & \textbf{73.8} & \textbf{55.3} \\
\midrule
\textbf{Qwen2.5-7B-Instruct} & 10.0 & 10.0 & 70.6 & 90.2 & 82.0 & 2.0 & 12.2 & 12.6 & 6.6 & 24.0 & 32.0 \\
\quad + TIR Prompting        & 6.7 & 10.0 & 68.2 & 64.6 & 78.2 & 15.5 & 14.8 & 18.3 & 9.5 & 23.6 & 31.0 \\
\quad + GRPO                 & 23.3 & \underline{26.7} & 78.0 & \textbf{92.8} & \underline{87.8} & 22.0 & \textbf{59.0} & \textbf{76.1} & \underline{30.6} & \underline{68.4} & \underline{56.5} \\
\quad + Reinforce ++         & \underline{26.7} & 23.3 & 78.0 & \underline{92.2} & \textbf{88.8} & \textbf{26.0} & 55.1 & \underline{68.9} & 25.2 & 64.9 & 54.9 \\
\quad + DAPO                 & 20.0 & 23.3 & \textbf{80.4} & 91.0 & \textbf{88.8} & \underline{24.0} & 57.7 & 68.4 & 28.6 & 65.5 & 54.8 \\
\rowcolor[RGB]{236,244,252} 
\quad + ARPO                 & \textbf{30.0} & \textbf{30.0} & \underline{78.8} & \underline{92.2} & \textbf{88.8} & \textbf{26.0} & \underline{58.8} & \textbf{76.1} & \textbf{31.1} & \textbf{71.5} & \textbf{58.3} \\

\bottomrule
\vspace{-4em}
\end{tabular}
\end{table*}

\subsection{Main Results}
\textbf{Results on Mathematical \& Knowledge-Intensive Reasoning.} Our main results are shown in Table~\ref{tab:main_table}. In a fair setting, ARPO consistently outperforms all trajectory-level RL algorithms, firmly establishing its superiority. Moreover, we highlight the following insights:

\begin{itemize}[leftmargin=1em]
\item \textbf{Ineffectiveness of Prompting Methods:} The Tool-integrated prompting (TIR) method~\citep{searcho1} fails to effectively explore superior tool-use behaviors. For both Qwen and Llama series models, performance improvements with TIR prompts are limited and even lower than direct reasoning. This suggests that relying solely on prompt engineering is insufficient for guiding LLMs toward optimal tool behaviors and may disrupt their inherent reasoning capabilities.

\item \textbf{Limitations of Trajectory-Level RL:} Compared to ARPO, three classic trajectory-level RL algorithms do not effectively harness the potential for tool-integrated reasoning. While the DAPO algorithm excels in single-turn reasoning tasks, it underperforms in multi-turn tool-call interaction, especially in knowledge-intensive scenarios. This aligns with our preliminary observations that trajectory-level RL algorithms struggle to stimulate step-level tool-use behavior learning in LLMs.

\item \textbf{Robust Performance of ARPO:} In the same experimental setup, ARPO consistently outperforms other RL algorithms across 10 datasets, achieving an average accuracy improvement of 4\% while maintaining competitive result on individual domains. Notably, it shows significant enhancements across different backbone models, including both Qwen and Llama series. These results underscore ARPO's efficiency, and strong adaptability across various model backbones and tasks.
\end{itemize}

\begin{table*}[!t]
\centering
\small
\caption{Overall performance on various deep search tasks, with accuracy results for each dataset obtained using llm-as-judge. The best results are indicated in \textbf{bold}, and the second-best results are \underline{underlined}. Results from larger or closed-source models are presented in \textcolor{gray!135}{gray} for reference.}
\label{tab:main_result}
\setlength\tabcolsep{0.5pt}
\fontsize{8.1pt}{10.5pt}
\selectfont
\begin{tabular}{
    p{2.8cm} % Method
    *{4}{>{\centering\arraybackslash}p{0.8cm}} % General AI Assistant
    *{4}{>{\centering\arraybackslash}p{0.8cm}}  % WebWalker (Easy, Med, Hard, Avg)
    *{4}{>{\centering\arraybackslash}p{0.75cm}} % Humanity's Last Exam
    >{\centering\arraybackslash}p{1.2cm}        % XBench
}
\toprule
\multirow{2}{*}{\textbf{Method}}
  & \multicolumn{4}{c}{\textbf{General AI Assistant}}
  & \multicolumn{4}{c}{\textbf{WebWalkerQA}}
  & \multicolumn{4}{c}{\textbf{Humanity's Last Exam}}
  & \textbf{XBench} \\
\cmidrule(lr){2-5} \cmidrule(lr){6-9} \cmidrule(lr){10-13} \cmidrule(lr){14-14}
  & Lv.1 & Lv.2 & Lv.3 & Avg.
  & Easy & Med. & Hard & Avg.
  & NS & CE & SF & Avg.
  & Avg. \\
\midrule

% 下面为正文数据部分，已按你的要求调整
\multicolumn{14}{l}{\textit{\textbf{Direct Reasoning (>=32B)}}} \\
Qwen3-32B-thinking & 26.2 & 12.1 & 0 & 14.9 & 6.9 & 1.1 & 2.9 & 3.1 & \underline{14.6} & \underline{9.8} & 8.4 & 12.6 & 14.0 \\
DeepSeek-R1-32B    & 21.5 & 13.6 & 0.0 & 14.2 & 7.5 & 1.4 & 4.2 & 3.8 & 6.6 & 5.1 & 6.5 & 6.4 & 10.0 \\
QwQ-32B            & 30.9 & 6.5  & 5.2 & 18.9 & 7.5 &2.1  &4.6  & 4.3 & 11.5 & 7.3 & 5.2 & 9.6 & 10.7 \\
GPT-4o             & \textcolor{gray!135}{23.1} & \textcolor{gray!135}{15.4} & \textcolor{gray!135}{8.3} & \textcolor{gray!135}{17.5} & \textcolor{gray!135}{6.7} &\textcolor{gray!135}{6.0}  &\textcolor{gray!135}{4.2}  & \textcolor{gray!135}{5.5} & \textcolor{gray!135}{2.7} & \textcolor{gray!135}{1.2} & \textcolor{gray!135}{3.2} & \textcolor{gray!135}{2.6} & \textcolor{gray!135}{18.0} \\
DeepSeek-R1-671B   & \textcolor{gray!135}{40.5} & \textcolor{gray!135}{21.2} & \textcolor{gray!135}{5.2} & \textcolor{gray!135}{25.2} & \textcolor{gray!135}{5.0} &\textcolor{gray!135}{11.8}  &\textcolor{gray!135}{11.3}  & \textcolor{gray!135}{10.0} & \textcolor{gray!135}{8.5} & \textcolor{gray!135}{8.1} & \textcolor{gray!135}{9.3} & \textcolor{gray!135}{8.6} & \textcolor{gray!135}{32.7} \\
o1-preview$^\dagger$ & \textcolor{gray!135}{-} & \textcolor{gray!135}{-} & \textcolor{gray!135}{-} & \textcolor{gray!135}{-} &\textcolor{gray!135}{11.9}  & \textcolor{gray!135}{10.4} &\textcolor{gray!135}{7.9}  & \textcolor{gray!135}{9.9} & \textcolor{gray!135}{12.9} & \textcolor{gray!135}{8.1} & \textcolor{gray!135}{6.6} & \textcolor{gray!135}{11.1} & \textcolor{gray!135}{-} \\
\midrule
\multicolumn{14}{l}{\textit{\textbf{Single-Enhanced Method (Qwen3-8B)}}} \\
Vanilla RAG  & 28.2 & 15.4 & \textbf{16.7} & 20.4 &  8.9 & 10.7 & 9.9 & 10.0 & 5.1 & 1.6 & \underline{12.9} & 5.8 & 8.0 \\
Search-o1    & 35.9 & 15.4 & 0.0  & 21.4 &  6.7 & 15.5 & 9.7 & 11.5 & \underline{7.6} & 2.7 & 5.3  & 6.4 & 10.0 \\
WebThinker   & 43.6 & 11.5 & 0.0  & 22.3 & 6.7 & 13.1 & 16.9 & 13.0 & 7.3 & \underline{4.0} & 6.3 & 6.6 & 13.0 \\
ReAct        & 35.9 & 17.3 & \underline{8.3}  & 23.3 &  8.9 & \underline{16.7} & 18.3 & 15.5 & 4.2 & \underline{4.0} & 6.3 & 4.6 & 16.0 \\
\multicolumn{14}{l}{\textit{\textbf{RL-based Method (Qwen3-8B)}}} \\
Qwen3-8B & 28.1 & 15.4 & \textbf{16.7} & 20.4 & 0.0 & 2.4 & 2.8 & 2.0 & 3.9 & 2.7 & 8.4  & 4.6& 9.0 \\
\;\; + GRPO & \underline{48.7} & \underline{25.0} & \underline{8.3} & \underline{32.0} & \underline{24.4} & \textbf{33.3} & \underline{26.8} & \underline{29.0} & \textbf{7.9} & \underline{4.0} & 10.5  & \underline{7.8} & \underline{20.0}\\
\rowcolor[RGB]{236,244,252}\;\; + ARPO & \textbf{53.9} & \textbf{32.7} & \textbf{16.7} & \textbf{38.8} & \textbf{26.7} & \textbf{33.3} & \textbf{29.6} & \textbf{30.5} & 7.3 & \textbf{6.7} & \textbf{15.8} & \textbf{8.8} &\textbf{25.0} \\

\midrule
\multicolumn{14}{l}{\textit{\textbf{Single-Enhanced Method (Qwen3-14B)}}} \\
Vanilla RAG  & 38.5 & 19.2 & \underline{8.3}  & 25.2 & 17.8 & 13.1 & 11.3 & 13.5 & 5.5 & 6.3 & 9.4 & 6.0 & 15.0 \\
Search-o1    & 48.7 & 23.1 & 0.0  & 30.1 &  11.1 & 21.4 & 16.9 & 17.5 & 6.4 & 4.0 & 10.5 & 6.8 & 21.0 \\
WebThinker   & 48.7 & 26.9 & \underline{8.3}  & 33.0 &  13.3 & 23.8 & 18.3 & 19.5 & 7.0 & 4.0 & 9.5 & 7.0 & 23.0 \\
ReAct        & 48.7 & 25.0 & \underline{8.3}  & 32.0 &  11.1 & 20.2 & 12.7 & 15.5 & 5.8  & 5.3 & 10.5 & 6.6 & 20.0 \\
\multicolumn{14}{l}{\textit{\textbf{RL-based Method (Qwen3-14B)}}} \\
Qwen3-14B & 33.3 & 13.5 & 0.0 & 19.4 & 6.7 & 2.4 & 4.2 & 4.0 & 5.5 & \underline{6.7} & 11.6 & 6.8 &14.0 \\
\;\; + GRPO & \underline{51.3} & \underline{34.6} & 0.0 & \underline{36.9} & \underline{28.9} & \underline{33.3} & \underline{26.8} & \underline{30.0} & \underline{7.9} & \underline{6.7} & \underline{12.6} & \underline{8.6} & \underline{27.0} \\
\rowcolor[RGB]{236,244,252}\;\; + ARPO & \textbf{56.4} & \textbf{40.4} & \textbf{16.7} & \textbf{43.7} & \textbf{31.1} & \textbf{42.9} & \textbf{31.0} & \textbf{36.0} & \textbf{10.3} & \textbf{10.7} & \textbf{13.7} & \textbf{10.0} &\textbf{32.0} \\
\bottomrule
\vspace{-3em}
\end{tabular}
\end{table*}

\textbf{Results on Deep Search Tasks.} To further verify the effectiveness of our ARPO in challenging deep search scenarios, we compare the performance of the Qwen3 series models, trained with only 1k RL samples, against a series of strong baseline methods. Our observations are as follows:

\begin{itemize}[leftmargin=1em]
\item \textbf{Generalization of ARPO in Deep Search Domain:} In deep search scenarios, even the most advanced LLMs like GPT-4o and DeepSeek-R1-671B achieve limited performance, scoring only 2\% and 8.6\% on the HLE benchmark respectively. In contrast, ARPO demonstrates exceptional performance using only the Qwen3-8 and 14B models, achieving pass@1 scores of 10.0\% and 43.2\% on the HLE and GAIA benchmarks. Notably, during the RL phase, ARPO is trained with just 1K samples from an open-source web search dataset, showcasing its efficiency in leveraging tool-integrated reasoning capabilities.

\item \textbf{Importance of Step-Level Tool Use Behavior Exploration:} ARPO consistently outperforms GRPO in both average performance and individual benchmarks, with a notable 6\% improvement on the GAIA and WebwalkerQA benchmarks. This highlights the importance of ARPO's algorithmic design, which balances global and step-level sampling. This balance promotes diverse behavior exploration by LLMs during high-entropy tool-use steps, crucial for deep search scenarios involving frequent tool invocation.
\end{itemize}

\subsection{Quantitative Analysis}

\begin{figure}[t]
    \centering
    \includegraphics[width=1\linewidth]{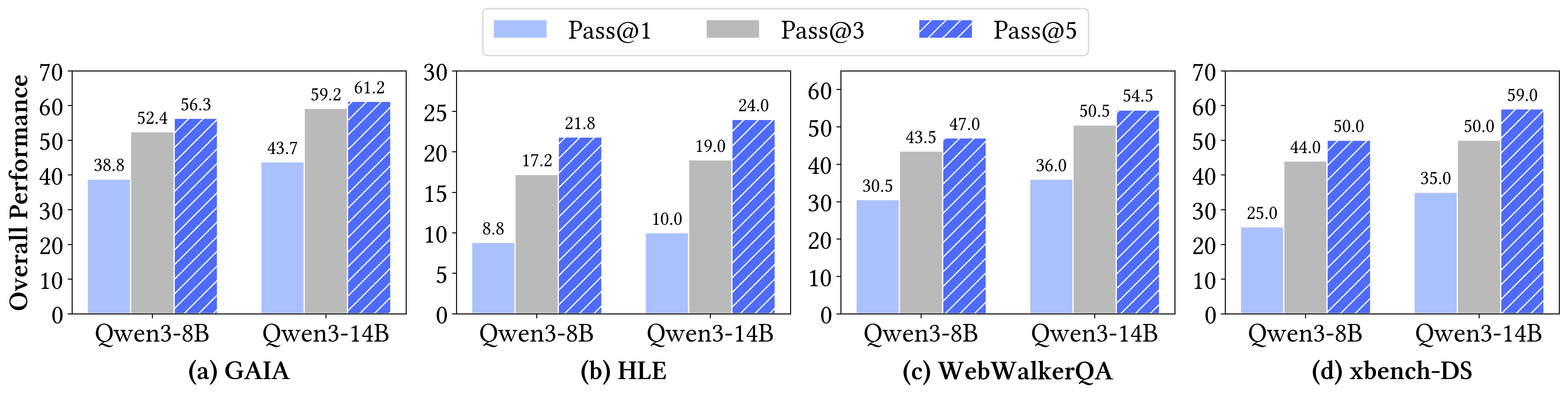}
    \vspace{-1em}
    \caption{Analysis of Qwen3-8B and Qwen3-14B using ARPO across Pass@1 to Pass@5 metrics.}
    \vspace{-1em}
    \label{fig:passk}
\end{figure}

\textbf{Analyzing Sampling at Scale.} Due to the dynamic and multi-round interaction characteristics of Deepsearch evaluation, Pass@1 is insufficient to capture the model's potential for tool usage. Consequently, we conducted further sampling analysis on Pass@3 and Pass@5, as illustrated in Figure \ref{fig:passk}. Both the 8B and 14B models demonstrated consistent improvements and a scaling trend in Pass@3 and Pass@5 following the ARPO alignment stage. Notably, our Qwen-14B with ARPO achieved remarkable performance on Pass@5, particularly with \textbf{GAIA at 61.2\%, HLE at 24.0\% and xbench-DR at 59\%.} This stable enhancement in Pass@K is primarily attributed to ARPO's ability to explore fine-grained tool-use behaviors more efficiently, thereby expanding the sampling space and achieving both inference efficiency and sampling diversity. \footnote{Since xbench-DR is composed entirely of Chinese questions, we utilize Chinese prompts to analyze the pass@k results, leading to a performance improvement compared to Table~\ref{tab:main_result}.}

\begin{wrapfigure}[12]{r}{0.46\linewidth}
\vspace{-0.5em} % 上方间距
    \centering
    \includegraphics[width=0.99\linewidth]{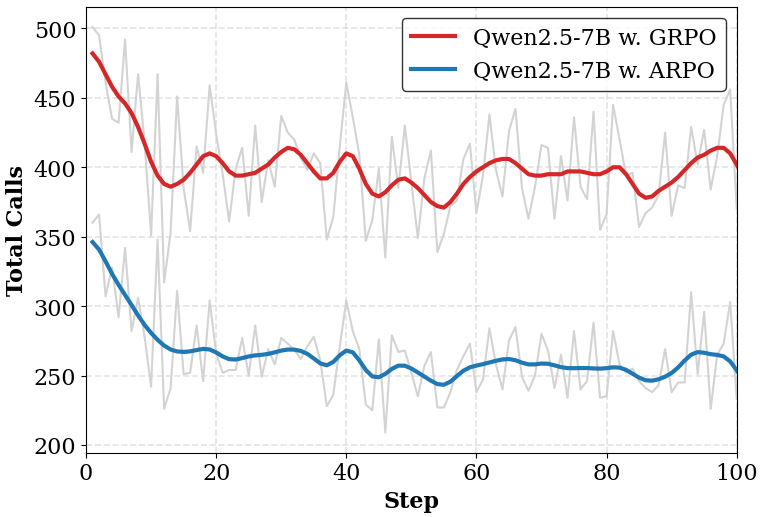}
    \vspace{-2em}
    \caption{Comparison of Tool-Call Efficiency for Qwen2.5-7B: GRPO vs. ARPO}
    %\vspace{1em} % 下方间距
    \label{fig:tool-call}
\end{wrapfigure}

\textbf{Tool-Call Efficiency Analysis.} In agentic RL training, increasing the number of tool calls often results in substantial financial  costs. Therefore, an effective agentic RL algorithm must ensure efficient tool usage. To assess the tool usage efficiency of ARPO during training, we compare it with GRPO on Qwen2.5-7B. As shown in Figure \ref{fig:tool-call}, ARPO achieves superior overall accuracy compared to GRPO while \textbf{using only half the number of tool calls.} This efficiency is attributed to ARPO's unique entropy-based adaptive rollout mechanism, which selectively explores branches only during high-entropy tool-call steps. This approach significantly expands the\\ exploration space for tool behavior while greatly\\ reducing the number of tool calls.

\textbf{Ablations of Browser Agents.} To further investigate the importance of the browser agent in the Deepsearch task, we designed three browser settings, ranked from weakest to strongest in terms of capability: \textbf{(1)} no browser with only snippets; \textbf{(2)} a browser agent with a similar scale to the reasoning model, and \textbf{(3)} a larger-parameter browser agent.

\begin{wraptable}{r}{0.5\textwidth}
\centering
\small
\vspace{-2em}
\caption{Ablation studies of the backbone model of browser agents in deep search tasks.}
\label{tab:ablation_broswer}
\setlength\tabcolsep{1.2pt} % Adjust column separation for this subtable
\renewcommand{\arraystretch}{1.1} % 增加行距，1.2表示增大为1.2倍
\begin{tabular}{lcccc}
\toprule
\textbf{Method} & \textbf{GAIA} & \textbf{HLE} & \textbf{WebWalk.}. & \textbf{Avg.} \\
\midrule
\multicolumn{5}{l}{\textit{\textbf{Qwen3-8B}}} \\
~~+ Snippet only       & 33.0 & 7.5 & 29.0 &  23.2 \\
~~+ Qwen3-8B Browser   & 38.8 & \textbf{8.8} & 30.5 &  26.0 \\
~~+ QWQ-32B Browser    & \textbf{38.8} & 8.2 & \textbf{33.0} &  \textbf{26.6} \\
\midrule
\multicolumn{5}{l}{\textit{\textbf{Qwen3-14B}}} \\
~~+ Snippet only       & 35.0 & 8.4 & 31.0 &  24.8 \\
~~+ Qwen3-14B Browser  & 43.7 & 10.0 & 36.0 & 29.9 \\
~~+ QWQ-32B Browser    & \textbf{47.6} & \textbf{32.3} & \textbf{38.4} & 39.4 \\

\bottomrule
\end{tabular}
\end{wraptable}
As shown in Table~\ref{tab:ablation_broswer}, results show that the setting without a browser exhibits the worst performance consistency, indicating that relying solely on rule-generated web snippet summaries is insufficient to provide the necessary information support in deep search tasks. This highlights the necessity of web content fetching and browsing. As the capability of the browser agent increases, model performance also improves significantly, demonstrating that a more powerful search agent can more effectively integrate information and extract key details relevant to the question. In summary, the capability of the external browser agent is highly correlated with the accuracy of the Deepsearch task and shows a clear upward trend as its scale increases.

\begin{figure}[t]
    \centering
    \includegraphics[width=1\linewidth]{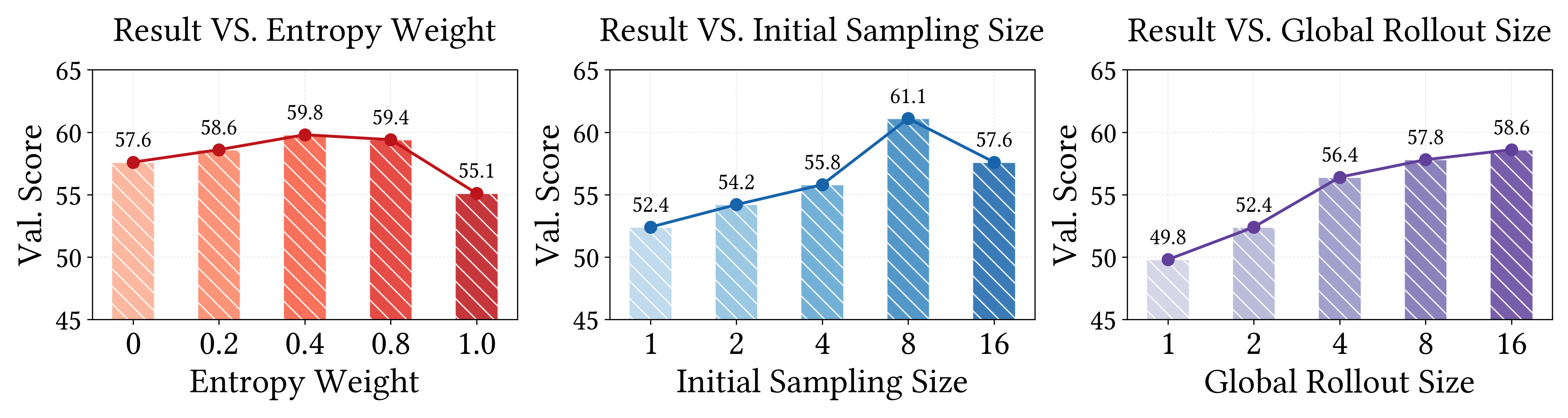}
    \vspace{-2em}
    \caption{Scaling analysis of different Hyper-parameters in Qwen2.5-7B with ARPO. The detailed setting can be found in Appendix~\ref{app:scaling_exp}.}
    \label{fig:scaling_exp}
\end{figure}

\subsection{Scaling Analysis of ARPO}
To verify the scalability of ARPO and gain deeper insights into its characteristics, we use the Qwen2.5-7B model as the backbone for a scaling analysis of three core parameters: \textit{entropy value}, \textit{global rollout size}, and \textit{initial sampling size}. Our observations are as follows:

\textbf{Entropy Value ($\Delta H_t$):} As shown in Figure~\ref{fig:scaling_exp} (left), model performance increases with rising entropy values, peaking at 0.4. This indicates that integrating a moderate amount of entropy as a clue for partial sampling substantially enhances the model's ability to explore rare tool-use behaviors, thereby improving training outcomes. However, as entropy reaches 1.0, performance declines, suggesting a trade-off in the weight of entropy in sampling. Over-reliance on entropy may reduce sampling diversity, confirming the necessity of balancing base sampling probabilities $\alpha$ with entropy in ARPO.

\textbf{Initial Sampling Size ($N$):} Figure~\ref{fig:scaling_exp} (middle) illustrates that as the initial sampling size increases, model performance improves, peaking at 8. Notably, with a global rollout size of 16, increasing the initial sampling size from 0 to 8 shifts the global-to-partial sampling ratio from 1:15 to 1:1. This underscores the importance of balancing sampling proportions for imrpoving performance. As anticipated, increasing the size to 16 results in a great performance decline. This is because it leads to complete global sampling, which disrupts the dynamic sampling balance.

\textbf{Global Rollout Size ($M$):} As depicted in the Figure~\ref{fig:scaling_exp} (right), increasing the global rollout size enhances model performance, indicating that the ARPO algorithm is scalable and can improve generalization performance with larger sizes.

\section{Related Work}

\textbf{Reinforcement Learning with Verifiable Reward.} Recently, Reinforcement Learning with Verifiable Rewards (RLVR)~\citep{tulu3,rlhf} has become a leading approach in Reinforcement Learning through Human Feedback (RLHF), particularly excelling in enhancing mathematical and programming reasoning~\citep{deepseekmath,deepseek-r1,qwen3,qwen2,qwen_qwq,qwen2.5blog,how_abilities,wemath}. OpenAI o1~\citep{2409_openai_o1} first showcased RL's effectiveness in large-scale reasoning tasks. Building on this, models like DeepSeek R1~\citep{deepseek-r1}, QwQ~\citep{qwq-32b-preview}, and Kimi k1.5~\citep{team2025kimi} aim to replicate and surpass its performance. To improve RL algorithms' performance and stability, researchers have developed models like DAPO~\citep{yuDAPO} and SimpleRLZoo~\citep{simplerlzoo}, exploring algorithm design across various RL modules\citep{2503-Open-Reasoner-Zero,vapo,Group-in-Group,noisyrollout,reloo1,reloo2,autoif,hu2025reinforce++}. \citeauthor{critical_token} identified key tokens affecting errors and showed that replacing them can alter model behavior. Studies~\citep{gandhi2025cognitive,LLM_can_easily_learn} found RLVR primarily learns format over content, while several works \citep{vassoyan2025ignore,20/80,zhaoxin_entropy,wang2025reinforcement} pointed out key tokens to high-entropy tokens to explore RL learning's essence. However, RLVR algorithms specifically for LLM agents remain underexplored. This paper uses entropy as a criterion to investigate reinforcement learning algorithms suited for LLM agent behavior.

\textbf{Agentic Reinforcement Learning.} Reinforcement learning (RL) is essential for enabling LLM agents to adapt to dynamic and open environments~\citep{AgentRewardBench,shridhar2020alfworld,GAIA}. Foundational works like DQN~\citep{dqn} and AlphaZero~\citep{alphazero} demonstrate that self-play-based RL can equip agents with capabilities from natural language understanding to strategic gameplay~\citep{textgame_rl}. Building on this, value-based RL approaches have been employed to enhance embodied intelligence in hardware control and complex gaming tasks~\citep{agentrl_2,agentrl_3,agent_rl_4,agent_rl_5,ppo,Advantage-Weighted-Regression}. Recent efforts, exemplified by RAGEN~\citep{ragen,zhou2024archer}, integrates reasoning states and environmental interactions into turn-level responses using trajectory-level RL. To improve tool-integrated reasoning, studies ~\citep{searchr1,retool,r1searcher,searchr1,chen2025research,retool,torl,zerosearch,webthinker,2505_ARTIST} employ rule-based RL to teach LLMs how to autonomously invoke external tools (e.g. search engines, Python compilers) to boost reasoning accuracy. Further research, including ToolRL~\citep{qian2025toolrl}, Tool-Star~\citep{dong2025toolstar}, and OTC~\citep{wang2025otc} explores multi-tool integration and tool-use efficiency. A series of works led by Kimi Deepresearcher~\footnote{\url{https://moonshotai.github.io/Kimi-Researcher/}} and Websailor~\citep{li2025websailor} optimize RL algorithms to better adapt to deepsearch's long context scenarios. While most works improve tool invocation through reward shaping and rollout mechanisms, simply applying trajectory-level RL fails to effectively capture the multi-turn, long-horizon characteristics of LLM-based agent behavior. This motivates the proposal of ARPO to attempt learning step-level tool-use behavior patterns.

\section{Conclusion}
In conclusion, we present Agentic Reinforced Policy Optimization (ARPO), an innovative reinforcement learning algorithm tailored for training multi-turn, LLM-based agents. Our experiments reveal that LLMs exhibit high token entropy after tool usage. ARPO leverages this by incorporating an entropy-based adaptive rollout mechanism, balancing global and step-level sampling to encourage diverse exploration in high-entropy tool-use phases. By integrating Advantage Attribution Estimation, ARPO enables LLMs to internalize advantage differences in stepwise tool-use interactions. Across 13 challenging benchmarks in computational reasoning, knowledge reasoning, and deep search domains, ARPO consistently outperforms traditional trajectory-level RL algorithms. Remarkably, it achieves great performance with only half the tool-use budget of other methods, offering a scalable solution for aligning LLM-based agents with dynamic environments.

\bibliography{iclr2025_conference}
\bibliographystyle{iclr2025_conference}
\clearpage

\appendix

\section*{Appendix}
\startcontents[sections]
\printcontents[sections]{l}{1}{\setcounter{tocdepth}{3}}

\section{Datasets}
\label{app:datasets}
In this section, we provide a detailed introduction to the datasets used in ARPO's experiments.
\subsection{Mathematical Reasoning Benchmarks}
\begin{itemize}[leftmargin=1em]
\item \textbf{AIME24~\footnote{\url{https://huggingface.co/datasets/HuggingFaceH4/aime_2024}}} is a dataset in evaluating the mathematical reasoning ability of models. It consists of 30 challenging math problems. All of them are from the American Invitational Mathematics Examination. The problems in the AIME24 dataset cover a wide variety of mathematical fields such as algebraic equations and geometric puzzles. Due to the difficulty characteristics and the richness of question types, it has become a popular benchmark for evaluating the reasoning performance of models, and is widely used in multiple related research experiments.

\item \textbf{AIME25~\footnote{\url{https://huggingface.co/datasets/math-ai/aime25}}} consists of 30 challenging math problems. It is directly composed of the real questions from the American Invitational Mathematics Examination (AIME I \& II) newly released in February 2025. AIME25's knowledge areas are extremely wide. It deeply covers core mathematical sections such as algebra, geometry, number theory, and combinatorial mathematics. This characteristic enables the AIME25 dataset to effectively distinguish the mathematical reasoning abilities of different models.

\item \textbf{MATH500~\citep{math500}} is selected by OpenAI from the MATH evaluation dataset. It contains 500 high-difficulty math problems. These problems cover multiple mathematical fields such as algebra, geometry, calculus, and number theory. The difficulty is close to or exceeds the college level. In academic research, MATH500 dataset is often used to evaluate the performance of various reasoning models. 

\item \textbf{MATH~\citep{MATH}} is a significant academic dataset. It is designed to test and enhance models' mathematical reasoning skills. It covers a wide range of mathematical fields, including abstract algebra, calculus, and discrete mathematics. The dataset divides training data into three levels, which helps effectively evaluate model performance at different stages. 

\item \textbf{GSM8K~\citep{cobbe2021gsm8k}} is an elementary school math problem dataset released by OpenAI. These problems require 2 to 8 steps to solve, mainly through a series of basic calculations to obtain the final answer. This dataset is primarily used to test the logical and mathematical abilities of models and has been applied in multiple benchmark tests.

\end{itemize}
\subsection{Knowledge-Intensive reasoning benchmarks}
\begin{itemize}[leftmargin=1em]
\item \textbf{HotPotQA~\citep{hotpotqa}} is a question-answering dataset for multi-hop. All the documents are sourced from Wikipedia, which provides the dataset with a rich knowledge base and relatively structured information. It is an important benchmark for evaluating the ability of LLMs to understand complex search tasks.

\item \textbf{2WikiMultihopQA~\citep{2wiki}} is a dataset specifically designed for the multi-hop question-answering task. It aims to test and evaluate the ability of natural language processing models to answer questions that require multi-step reasoning and the integration of information from different documents.

\item \textbf{Musique~\citep{musique}} is a question-answering dataset specifically designed for the multi-hop question-answering task. Musique aims to be a challenging benchmark for evaluating models' multi-hop question-answering ability. It promotes the development of models from simple information retrieval to deeper semantic understanding and logical reasoning.
\end{itemize}
\subsection{Deep Search Benchmarks}
\begin{itemize}[leftmargin=1em]
\item \textbf{GAIA~\citep{GAIA}} is designed to evaluate the comprehensive capabilities of LLMs in real world tasks. The dataset contains 466 carefully designed questions to test the performance of AI systems in basic capabilities. It contains tasks such as reasoning, web browsing, and tool use. The proposal of GAIA provides a new framework for the evaluation of general artificial intelligence assistants.

\item \textbf{HLE~\citep{HLE}} is an emerging and highly challenging benchmark dataset. It aims to deeply evaluate the performance of LLMs when faced with complex questions requiring deep understanding and complex reasoning. This dataset covers a large number of marginal, interdisciplinary problems that demand highly abstract thinking to solve. Different from traditional benchmarks, HLE aims to simulate an ultimate test of AI intelligence. 

\item \textbf{WebWalker~\citep{2501_WebWalker}} is a dataset used to evaluate the performance of LLMs in web traversal tasks. This dataset contains 680 question-answer pairs. It aims to address the limitations of LLMs when dealing with complex information. Additionally, it improves the models' capabilities in multi-hop reasoning and dynamic web page structures.

% \item \textbf{SimpleQA~\citep{simpleqa}} is a classic Knowledge Base Question Answering (KBQA) dataset. It is designed to evaluate systems' ability to retrieve facts from large knowledge bases. This ability serves to answer simple natural language questions. It helps researchers evaluate the performance of models in entity linking and relation extraction.

\item \textbf{xbench-DeepSearch~\citep{chen2025xbench}} is an evaluation set for assessing the deep search capabilities of AI agents. This dataset takes full consideration of the breadth of the search space and the depth of reasoning. Different from existing knowledge search benchmarks, xbench-DeepSearch is more capable of examining the high-order capabilities of agents.
\end{itemize}

\section{Baselines}
\label{app:baselines}
In this section, we introduce baselines used in our work.
\subsection{Direct Reasoning}
\begin{itemize}[leftmargin=1em]
\item \textbf{Qwen2.5 Series~\citep{qwen2.5}} is a series of LLMs developed by the Alibaba team. It includes the general-purpose language model Qwen2.5, the programming-specific model Qwen2.5-Coder, and the mathematics-specific model Qwen2.5-Math. The Qwen2.5 series of models have been pretrained on large scale datasets. Compared with past Qwen series of models, the Qwen2.5 series have a richer knowledge reserve. In addition, it has good performance in various tasks such as programming, mathematics, and instruction following.

\item \textbf{Llama3.1 Series~\citep{llama3}} is a series of natural language generation model launched by Meta. It includes three specifications: 8B, 70B, and 405B. These models can handle longer text inputs and generate more coherent long-text outputs. This series of models also performs well in multilingual tasks. The Llama 3.1 series of models have undergone performance tests on more than 150 benchmark datasets. Its large scale model is competitive with leading base models in a series of tasks. The smaller 8B and 70B models also perform well in comparisons with closed-source and open-source models with a similar number of parameters.

\item \textbf{Qwen 3 Series~\citep{qwen3}} is a series of open-source model developed by Alibaba. The Qwen3 series of models includes 2 Mixture-of-Experts (MoE) models and 6 Dense models, with the number of parameters ranging from 0.6B to 235B. Qwen3 natively supports the thinking mode and non-thinking mode. In the thinking mode, the model reasons step by step and is suitable for handling complex problems. The non-thinking mode can provide a fast, nearly instant response and is suitable for simple problems. Qwen3 builds a training corpus based on approximately 36 trillion tokens, ensuring the model's powerful capabilities and flexibility.

\item \textbf{QwQ~\citep{qwen_qwq}} is an open-source inference model launched by Alibaba team. It focuses on enhancing AI's capabilities in mathematics, programming, and complex logical reasoning. The QwQ-32B is a dense model with 32 billion parameters. It surpasses most existing models in core tasks such as mathematical reasoning and code-generation ability. The QwQ-32B achieves breakthroughs through innovative multi-stage reinforcement learning. Its core training approach lies in gradually expanding general capabilities while consolidating specialized advantages.

\item \textbf{DeepSeek-R1~\citep{deepseek-r1}} is a reasoning model developed by DeepSeek-AI. DeepSeek-R1 is trained using reinforcement learning. The inference process involves a large amount of reflection and verification, and the length of the thought chain can reach tens of thousands of tokens. It performs outstandingly in tasks such as mathematics, code, and various complex logical reasoning.

\item \textbf{GPT-4o~\citep{gpt_4o_system_card}} is a multimodal LLM released by OpenAI. GPT-4o can accept any combination of text, audio, and images as input. In addition, it can generate any combination of text, audio, and images as output. GPT-4o has achieved performance comparable to that of GPT-4 Turbo in aspects such as text, reasoning, and coding. Moreover, it has also set new highs in performance scores for multilingual, audio, and visual functions. 

\item \textbf{o1-preview~\citep{gpt_4o_system_card}} is a preview version model in the o1 series of LLMs launched by OpenAI. It represents an important breakthrough in the field of reasoning. o1-preview is based on the GPT-4 architecture and trained through reinforcement learning. It aims to enhance the reasoning ability for complex tasks and the ability to solve practical problems. It can exhibit powerful abilities in tasks that require in-depth reasoning.
\end{itemize}
\subsection{Trajectory-level RL Algorithms}
\begin{itemize}[leftmargin=1em]
\item \textbf{GRPO~\citep{deepseekmath}} is a reinforcement learning algorithm based on policy optimization. It aims to address the balance issue among stability, sample efficiency, and theoretical guarantees in traditional policy optimization methods. By introducing the concept of relative advantage, it simplifies the calculation while maintaining the theoretical guarantee of policy improvement. The GRPO algorithm is applicable to reinforcement learning tasks in both continuous and discrete action spaces.

\item \textbf{DAPO~\citep{yuDAPO}} is a LLM reinforcement learning algorithm developed by ByteDance Labs. It aims to address the key challenges of large-scale RL training. It performs outstandingly in complex tasks such as mathematical reasoning and code generation. The Clip-Higher strategy proposed by DAPO effectively increases the entropy value, facilitating the generation of more diverse samples. In addition, it introduces mechanisms such as dynamic sampling, Token-Level Policy Gradient Loss calculation, and Overlong Reward Shaping to stabilize the training process. 

\item \textbf{REINFORCE++~\citep{hu2025reinforce++}} is a new algorithm for improved versions of the classic REINFORCE algorithm. Its core objective is to address the limitations of the original REINFORCE, and enhance performance by integrating multiple optimization strategies. REINFORCE++ typically incorporates a baseline function to reduce variance by subtracting the baseline. Through the baseline and TD estimation, REINFORCE++ makes the gradient update more stable. It doesn't need to wait for a complete trajectory and supports incremental updates. In addition, it avoids premature policy rigidity through entropy regularization.
\end{itemize}
\subsection{LLM-based Search Agent}
\begin{itemize}[leftmargin=1em]
\item \textbf{RAG~\citep{rag_lewis}} (Retrieval-Augmented Generation) is a technical approach that combines information retrieval with generative model. It aims to improve the accuracy, reliability, and timeliness of the output of generative models. Its core idea is: before generating an answer, first retrieve information related to the question from an external knowledge base, and then let the model generate a response based on the retrieved content. This can solve the problem of internal knowledge deficiency or hallucination within the model to some extent.

\item \textbf{Search-o1~\citep{searcho1}} is an Agentic search-enhanced reasoning model framework. It is mainly designed to address the knowledge deficiency problem existing in the reasoning process. By integrating the Agentic RAG mechanism and the Reason-in-Documents module, it improves the accuracy, coherence, and reliability of model reasoning. Experiments show that Search-o1 outperforms native reasoning and traditional RAG methods in complex reasoning tasks.

\item \textbf{WebThinker~\citep{webthinker}} is an open-source in-depth research framework launched by Renmin University of China. It endows LRMs with the ability to autonomously search, deeply explore web pages, and write research reports. WebThinker has developed a training strategy based on direct preference optimization. It uses training with preference data through iterative synthesis tools to enhance the tool utilization ability of LRMs.

\item \textbf{ReAct~\citep{yao2022react}} is an artificial intelligence method that combines reasoning and acting. It aims to enable models to solve complex tasks more effectively through a ``thinking while doing'' mode similar to human thinking. Its core idea is to break the limitation of the traditional model, allowing the model to actively generate reasoning steps and call external tools (such as search engines, databases, etc.) during the decision-making process, and finally obtain the answer through iterative optimization.
\end{itemize}

\section{Implementation Details}
\label{app:implementation}
\subsection{Supervised Fine-Tuning} 
As mentioned in Section~\ref{sec:training_guide}, during the supervised fine-tuning phase, we train the Qwen2.5-3B-Instruct model using the Llama Factory framework with a learning rate of \(7 \times 10^{-6}\). We employ DeepSpeed ZeRO-3~\citep{deepspeed} and FlashAttention2~\citep{flashattention} for optimization. The batch size is set to 128, with a weight decay of 0.1, and the model is trained for 3 epochs. We use BF16 mixed precision with a maximum input length of 4096 tokens. 

\subsection{Reinforcement Learning} In the ARPO phase, we implement the ARPO algorithm based on the VERL framework~\citep{sheng2024hybridflow}. Notably, all tool invocation results are excluded from loss calculation to prevent bias towards tool outputs. The loss computation only considers tokens involved in text reasoning and tool requests. We differentiate settings for Deep Reasoning Tasks and Deep Search Tasks:

\textbf{1. Deep Reasoning Tasks:} For models with 7B parameters, whether using ARPO or other trajectory-level RL methods, our standard setup includes a total training batch size of 128, a PPO mini-batch size of 16, a global rollout size of 16, and an initial sampling size of 8. Each interaction response length is capped at 4096 tokens. For ARPO rollouts, we set the entropy weight to 0.2, the parameter \(a\) to 0.5, and the threshold to 0.5. To stabilize training, the KL divergence coefficient in GRPO is set to 0. The reinforcement learning phase spans 2 epochs, conducted on 8 NVIDIA H800 GPUs.

\textbf{2. Deep Search Tasks:} For models with 8B parameters, we maintain the same settings as in the Deep Reasoning Tasks, except that each interaction response length is extended to 8192 tokens. For 14B models, the same parameters are used, but experiments are conducted on 16 NVIDIA H800 GPUs. Due to a limited dataset of 1K samples, the reinforcement learning phase lasts for 5 epochs.

\subsection{Details of Search}
During the training and testing phases, we used the Bing Web Search API as the retriever, configured with the US-English (US-EN) locale. Following a series of related works on RAG~\citep{flashrag,retrollm,dparag,followrag,2412_AR_MCTS}, we retrieved 10 web pages as supporting documents for each query.

For mathematical and knowledge reasoning, we evaluated using only the top 10 snippets. However, for deep search tasks, we fetched each page with up to 6000 tokens from the URLs and used a model of the same size as the reasoning model as a browser agent to refine the information.

\subsection{Scaling Experiment Setup} 
\label{app:scaling_exp}
In our scaling experiments, we align with the aforementioned settings: a total training batch size of 128, a PPO mini-batch size of 16, a global rollout size of 16, and an initial sampling size of 8. For ARPO rollouts, the entropy weight is 0.2, \(a\) is 0.5, and the threshold is 0.5. We vary specific parameters for targeted experiments while keeping others constant.

\subsection{C.3} 

\section{Theoretical Analysis and Proofs}
\subsection{Theoretical Analysis of Soft Advantage Estimation}
\label{app:Theoretical_proof}
In this section, we conduct a detailed theoretical analysis of Soft Advantage Estimation. First, we present the classic GRPO optimization objective:

\begin{small}
\begin{equation}
\begin{split}
J_{\text{GRPO}}(\theta) = \mathbb{E}_{(q,a) \sim D, \{o_i\}_{i=1}^G \sim \pi_{\theta_{\text{old}}}(\cdot \mid q)} \Bigg[ \frac{1}{G} \sum_{i=1}^G \frac{1}{|o_i|} \sum_{t=1}^{|o_i|} \min \Big( r_{i,t}(\theta) \hat{A}_{i,t}, \\
& \hspace*{-10.5em} \text{clip} \left( r_{i,t}(\theta), 1 - \epsilon, 1 + \epsilon \right) \hat{A}_{i,t} \Big) - \beta D_{\text{KL}}(\pi_{\theta} \parallel \pi_{\text{ref}}) \Bigg]
\end{split}
\label{eq:app grpo}
\end{equation}
\end{small}

For each individual problem, we define the optimization objective as:

\begin{small}
\begin{equation}
\begin{split}
J_{\text{GRPO}}^{q}(\theta) =  \frac{1}{G} \sum_{i=1}^G \frac{1}{|o_i|} \sum_{t=1}^{|o_i|} \min \Big( r_{i,t}(\theta), \text{clip} \left( r_{i,t}(\theta), 1 \pm \epsilon \right) \Big) \hat{A}_{i,t} - \beta D_{\text{KL}}(\pi_{\theta} \parallel \pi_{\text{ref}}) 
\end{split}
\label{eq:individual grpo}
\end{equation}
\end{small}

Therefore, the classical GRPO optimization objective can be expressed as:

\begin{small}
\begin{equation}
\begin{split}
J_{\text{GRPO}}(\theta) = \mathbb{E}_{(q,a) \sim D, \{o_i\}_{i=1}^G \sim \pi_{\theta_{\text{old}}}(\cdot \mid q)} [ J_{\text{GRPO}}^{q}(\theta) ]
\end{split}
\label{eq:classical grpo}
\end{equation}
\end{small}

Subsequently, we focus on analyzing $J_{\text{GRPO}}^{q}(\theta)$. Assume that for the inference of problem $q$, the partial rollout operation starts from the $l$-th token. We define two importance sampling ratios:

\begin{small}
\begin{equation}
\begin{split}
r_{i,t}(\theta)^{<l}= \frac{\pi_\theta(y_{i,t} \mid x, y_{i,<t})}{\pi_{\text{ref}}(y_{i,t} \mid x, y_{i,<t})},
\end{split}
\label{eq:app importance sampling ratio before l}
\end{equation}
\end{small}

\begin{small}
\begin{equation}
\begin{split}
r_{i,t}(\theta)^{>l}= \frac{\pi_\theta(p \mid x, q)}{\pi_{\text{ref}}(p \mid x, q)} · \frac{\pi_\theta(y_{i,t} \mid x, q, p, y_{i,<t})}{\pi_{\text{ref}}(y_{i,t} \mid x, q, p, y_{i,<t})},
\end{split}
\label{eq:app importance sampling ratio after l}
\end{equation}
\end{small}
where $r_{i,t}(\theta)^{<l}$ and $r_{i,t}(\theta)^{>l}$ represent the importance sampling ratios before and after the $l$-th token, respectively, $q$ represents the input question, $p$ represents shared tokens, and $y_{i,<t}$ in \ref{eq:app importance sampling ratio after l} represents the sequence from shared tokens to before the $t$-th token. In addition, we define $o_l^i$ as the $l$-th token of the $i$-th sequence. Then $J_{\text{GRPO}}^{q}(\theta)$ can be expressed as:

\begin{small}
\begin{equation}
\begin{split}
J_{\text{GRPO}}(\theta) &= \frac{1}{G} \sum_{i=1}^G \frac{1}{|o_i|} \Bigg[ \sum_{t=1}^{|o_l^i|} \min \Big( r_{i,t}(\theta)^{<l}, \text{clip} \left( r_{i,t}^{<l}(\theta), 1 \pm \epsilon \right) \Big) \hat{A}_{i,t} \\
& \hspace*{8.5em} + \sum_{t=|o_l^i|}^{|o_i|} \min \Big( r_{i,t}^{>l}(\theta), \text{clip} \left( r_{i,t}^{>l}(\theta), 1 \pm \epsilon \right) \Big) \hat{A}_{i,t} \Bigg] - \beta D_{\text{KL}}(\pi_{\theta} \parallel \pi_{\text{ref}}) \\
&= \frac{1}{G} \sum_{i=1}^G \frac{1}{|o_i|} \Bigg[ \sum_{t=1}^{|o_l^i|} \min \Big( r_{i,t}(\theta)^{<l}, \text{clip} \left( r_{i,t}^{<l}(\theta), 1 \pm \epsilon \right) \Big) \hat{A}_{i,t} \Bigg]\\
& \hspace*{4.5em} + \frac{1}{G} \sum_{i=1}^G \frac{1}{|o_i|} \Bigg[ \sum_{t=|o_l^i|}^{|o_i|} \min \Big( r_{i,t}^{>l}(\theta), \text{clip} \left( r_{i,t}^{>l}(\theta), 1 \pm \epsilon \right) \Big) \hat{A}_{i,t} \Bigg] - \beta D_{\text{KL}}(\pi_{\theta} \parallel \pi_{\text{ref}})\\
&= \frac{1}{G} \sum_{i=1}^G \frac{1}{|o_i|} \Bigg[ \sum_{t=1}^{|o_l^i|} \min \Big( r_{i,t}(\theta)^{<l}, \text{clip} \left( r_{i,t}^{<l}(\theta), 1 \pm \epsilon \right) \Big) \hat{A}_{i,t} \Bigg]\\
& \hspace*{0.5em} + \frac{1}{G} \sum_{i=1}^G \frac{|o_{l:i}|}{|o_i|} \cdot \frac{1}{|o_{l:i}|} \Bigg[ \sum_{t=|o_l^i|}^{|o_i|} \min \Big( r_{i,t}^{>l}(\theta), \text{clip} \left( r_{i,t}^{>l}(\theta), 1 \pm \epsilon \right) \Big) \hat{A}_{i,t} \Bigg] - \beta D_{\text{KL}}(\pi_{\theta} \parallel \pi_{\text{ref}})\\
&= \frac{1}{G} \sum_{i=1}^G \frac{1}{|o_i|} \Bigg[ \sum_{t=1}^{|o_l^i|} \min \Big( r_{i,t}(\theta)^{<l}, \text{clip} \left( r_{i,t}^{<l}(\theta), 1 \pm \epsilon \right) \Big) \hat{A}_{i,t} \Bigg]\\
& \hspace*{0.5em} + \frac{1}{G} \sum_{i=1}^G \frac{|o_i|-|o_l|}{|o_i| \cdot |o_{l:i}|} \Bigg[ \sum_{t=|o_l^i|}^{|o_i|} \min \Big( r_{i,t}^{>l}(\theta), \text{clip} \left( r_{i,t}^{>l}(\theta), 1 \pm \epsilon \right) \Big) \hat{A}_{i,t} \Bigg] - \beta D_{\text{KL}}(\pi_{\theta} \parallel \pi_{\text{ref}})\\
&= \frac{1}{G} \sum_{i=1}^G \frac{1}{|o_i|} \Bigg[ \sum_{t=1}^{|o_l^i|} \min \Big( r_{i,t}(\theta)^{<l}, \text{clip} \left( r_{i,t}^{<l}(\theta), 1 \pm \epsilon \right) \Big) \hat{A}_{i,t} \Bigg]\\
& \hspace*{6.5em} + \frac{1}{G} \sum_{i=1}^G \frac{1}{|o_{l:i}|} \Bigg[ \sum_{t=|o_l^i|}^{|o_i|} \min \Big( r_{i,t}^{>l}(\theta), \text{clip} \left( r_{i,t}^{>l}(\theta), 1 \pm \epsilon \right) \Big) \hat{A}_{i,t} \Bigg] \\
& \hspace*{0.5em} - \frac{1}{G} \sum_{i=1}^G \frac{|o_l|}{|o_i| \cdot |o_{l:i}|} \Bigg[ \sum_{t=|o_l^i|}^{|o_i|} \min \Big( r_{i,t}^{>l}(\theta), \text{clip} \left( r_{i,t}^{>l}(\theta), 1 \pm \epsilon \right) \Big) \hat{A}_{i,t} \Bigg] - \beta D_{\text{KL}}(\pi_{\theta} \parallel \pi_{\text{ref}})\\
&= \frac{1}{G} \sum_{i=1}^G \frac{|o_l|}{|o_i|} \Bigg[ \frac{1}{|o_l|} \sum_{t=1}^{|o_l^i|} \min \Big( r_{i,t}(\theta)^{<l}, \text{clip} \left( r_{i,t}^{<l}(\theta), 1 \pm \epsilon \right) \Big) \hat{A}_{i,t} \\
& \hspace*{6.5em} - \frac{1}{|o_{l:i}|}\sum_{t=|o_l^i|}^{|o_i|} \min \Big( r_{i,t}^{>l}(\theta), \text{clip} \left( r_{i,t}^{>l}(\theta), 1 \pm \epsilon \right) \Big) \hat{A}_{i,t} \Bigg]\\
& \hspace*{2.5em} + \frac{1}{G} \sum_{i=1}^G \frac{1}{|o_{l:i}|} \Bigg[ \sum_{t=|o_l^i|}^{|o_i|} \min \Big( r_{i,t}^{>l}(\theta), \text{clip} \left( r_{i,t}^{>l}(\theta), 1 \pm \epsilon \right) \Big) \hat{A}_{i,t} \Bigg] - \beta D_{\text{KL}}(\pi_{\theta} \parallel \pi_{\text{ref}})\\
\end{split}
\label{eq:app grpo before l after l}
\end{equation}
\end{small}

We make the following definitions:

\begin{small}
\begin{equation}
\begin{split}
J_i^{<l} = \frac{1}{|o_l|} \sum_{t=1}^{|o_l^i|} \min \Big( r_{i,t}(\theta)^{<l}, \text{clip} \left( r_{i,t}^{<l}(\theta), 1 \pm \epsilon \right) \Big) \hat{A}_{i,t},
\end{split}
\label{eq:app individual sequence grpo before l}
\end{equation}
\end{small}

\begin{small}
\begin{equation}
\begin{split}
J_i^{>l} = \frac{1}{|o_{l:i}|}\sum_{t=|o_l^i|}^{|o_i|} \min \Big( r_{i,t}^{>l}(\theta), \text{clip} \left( r_{i,t}^{>l}(\theta), 1 \pm \epsilon \right) \Big) \hat{A}_{i,t},
\end{split}
\label{eq:app individual sequence grpo after l}
\end{equation}
\end{small}

\begin{small}
\begin{equation}
\begin{split}
J_{\text{GRPO}}^{>l} = \frac{1}{G} \sum_{i=1}^G \frac{1}{|o_{l:i}|} \Bigg[ \sum_{t=|o_l^i|}^{|o_i|} \min \Big( r_{i,t}^{>l}(\theta), \text{clip} \left( r_{i,t}^{>l}(\theta), 1 \pm \epsilon \right) \Big) \hat{A}_{i,t} \Bigg],
\end{split}
\label{eq:app grpo after rollout}
\end{equation}
\end{small}
where $J_i^{<l}$ represents the optimization objective of the shared tokens part of the $i$-th chain, $J_i^{>l}$ represents the optimization objective after partial rollout of the $i$-th chain, 
and $J_{\text{GRPO}}^{>l}$ represents the optimization objective of directly performing the classical GRPO sampling operation starting from the $l$-th position. Then, the original optimization objective $J_{\text{GRPO}}(\theta)$ can be expressed as:

\begin{small}
\begin{equation}
\begin{split}
J_{\text{GRPO}}(\theta) = \frac{1}{G} \sum_{i=1}^G \frac{|o_l|}{|o_i|} [J_i^{<l} - J_i^{>l} ] + J_{\text{GRPO}}^{>l} - \beta D_{\text{KL}}(\pi_{\theta} \parallel \pi_{\text{ref}})
\end{split}
\label{eq:app grpo after rollout2}
\end{equation}
\end{small}

In this case, GRPO optimization objective under Soft Advantage Estimation can be expressed as the sum of a standard GRPO objective from the partial rollout position, and the weighted difference between the objectives before and after that point. The weight of each difference is closely related to the length of its corresponding reasoning chain.

\subsection{Theoretical Proof of GPG Theorem}
\label{app:Theoretical_proof_GPG}
\subsubsection{Transformer-based Policy}
The Transformer-based policy $\pi_{\theta}(a_t|s_t)$, by applying the chain rules, we have the following:
\begin{equation}
\begin{aligned}
    & \pi_{\theta}(OT_1 \;\;|\;\; IT_1, IT_2, ..., IT_{|input|}) \times \\
    & \pi_{\theta}(OT_2 \;\;|\;\; IT_1, IT_2, ..., IT_{|input|}, OT_1) \times \\
    & \pi_{\theta}(OT_3 \;\;|\;\; IT_1, IT_2, ..., IT_{|input|}, OT_1, OT_2) \times \\
    & ... \\
    & \pi_{\theta}(OT_{|output|} \;\;|\;\; IT_1, ..., IT_{|input|}, OT_1, ..., OT_{|output|-1}) \\
    = & \pi_{\theta}(OT_1, OT_2, ..., OT_{|output|} \;\;|\;\; IT_1, IT_2, ..., IT_{|input|}) \\
    = & \pi_{\theta}(MA \;\;|\;\; MS_1)
\end{aligned}
\label{equ:MacroAction1}
\end{equation}
where $IT_i$ and $OT_i$ are input tokens and output tokens, respectively; $MS_1 \triangleq < IT_1, IT_2, ..., IT_{|input|} >$ and $MA \triangleq < OT_1, OT_2, ..., OT_{|output|} >$ can be token as the \textbf{macro state} and the \textbf{macro action}, respectively.

In a more general form, we can split the complete output $OT_1, OT_2, ..., OT_{|output|}$ into $K$ segments, and get the generalized macro states and macro actions, i.e., $MS_i \triangleq < MS_{i-1}, MA_{i-1} > $ and $MA_i \triangleq < OT_m, OT_{m+1}, ..., OT_{m+n} >$. In this case, we have the following:
\begin{equation}
\begin{aligned}
    & \pi_{\theta}(MA \;\;|\;\; MS_1) \\
    = & \pi_{\theta}(MA_1 \;\;|\;\; MS_1) \times \\ 
    & \pi_{\theta}(MA_2 \;\;|\;\; MS_1, MA_1) \times \\
    & ... \\
    & \pi_{\theta}(MA_K \;\;|\;\; MS_1, MA_1, MA_2, ..., , MA_{K-1}) \\
   = & \pi_{\theta}(MA_1 \;\;|\;\; MS_1) \times \\ 
    & \pi_{\theta}(MA_2 \;\;|\;\; MS_2) \times \\
    & ... \\
    & \pi_{\theta}(MA_K \;\;|\;\; MS_K) \\ 
   = & \prod_{T=1}^{K} \pi_{\theta}(MA_T \;\;|\;\; MS_T)
\end{aligned}
\label{equ:MacroAction2}
\end{equation}
where $T$ represents the \textbf{macro timestep}.

\subsubsection{Derivation of the GPG Theorem}
Given the macro states and macro actions defined above, we can get the Generalized Policy Gradient Theorem (for Transformer-based policies):
\begin{equation}
\begin{aligned}
    \nabla_{\theta} J(\theta) 
    = & \mathbb{E}_{\tau \sim \pi_\theta}\{ \sum_{T=1}^{K} [\nabla_{\theta}\log\pi_{\theta}(MA_T|MS_T) \Phi_T ] \}
\end{aligned}
\end{equation}

\textbf{A key advantage of the GPG Theorem is that it allows macro-action segmentation of arbitrary length}. This flexibility makes the theorem highly practical: for instance, enabling trajectory splitting based on special tokens.

The proof is as follows:
\begin{align}
    & \nabla_{\theta} J(\theta)  \\
    = & \nabla_{\theta} \mathbb{E}_{\tau \sim \pi_\theta}[R(\tau)] \\
    = & \nabla_{\theta} \sum_{\tau}P(\tau;\theta)R(\tau) \\
    = & \sum_{\tau} \nabla_{\theta} P(\tau;\theta)R(\tau) \\
    = & \sum_{\tau} P(\tau;\theta) \frac{\nabla_{\theta} P(\tau;\theta)}{P(\tau;\theta)} R(\tau) \\
    = & \sum_{\tau} P(\tau;\theta) \nabla_{\theta} \log P(\tau;\theta) R(\tau) \\
    = & \sum_{\tau} P(\tau;\theta) \nabla_{\theta} [ \log \mu(s_1) \prod_{t=1}^{H} \pi_{\theta}(a_t|s_t) P(s_{t+1}|s_t,a_t) ] R(\tau) \\
    = & \sum_{\tau} P(\tau;\theta) \nabla_{\theta} [ \log \prod_{t=1}^{H} \pi_{\theta}(a_t|s_t) P(s_{t+1}|s_t,a_t) ] R(\tau) \label{equ:transition1} \\
    = & \sum_{\tau} P(\tau;\theta) \nabla_{\theta} [ \log \prod_{t=1}^{H} \pi_{\theta}(a_t|s_t)] R(\tau) \label{equ:transition2} \\
     = & \sum_{\tau} P(\tau;\theta) \nabla_{\theta} [\log \prod_{T=1}^{K} \pi_{\theta}(MA_T|MS_T) ] R(\tau) \label{equ:transition3}  \\
    = & \sum_{\tau}P(\tau;\theta) [\sum_{T=1}^{K}\nabla_{\theta}\log\pi_{\theta}(MA_T|MS_T) ] R(\tau) \\
    = & \sum_{\tau}P(\tau;\theta) [\sum_{T=1}^{K}\nabla_{\theta}\log\pi_{\theta}(MA_T|MS_T) R(\tau)] \\
    = & \mathbb{E}_{\tau \sim \pi_\theta}\{ \sum_{T=1}^{K} [\nabla_{\theta}\log\pi_{\theta}(MA_T|MS_T) R(\tau)] \} \label{equ:GPG1} \\
    = & \mathbb{E}_{\tau \sim \pi_\theta}\{ \sum_{T=1}^{K} [\nabla_{\theta}\log\pi_{\theta}(MA_T|MS_T) \Phi_T] \} \label{equ:GPG2}
\end{align}

Key steps in the proof are presented below:

1. From Equation (\ref{equ:transition1}) to Equation (\ref{equ:transition2}), this is because $s_{t+1} = [s_t, a_t]$ for Transformer-based policy, so we have $P(s_{t+1}|s_t,a_t) = 1$. 

2. From Equation (\ref{equ:transition2}) to Equation (\ref{equ:transition3}), this is because $s_{t+1} = [s_t, a_t]$ for Transformer-based policy, so we have the following derivation: 
\begin{equation}
\begin{aligned}
    & \prod_{t=1}^{H} \pi_{\theta}(a_t|s_t) \\
    = & \pi_{\theta}(a_1|s_1) \times \pi_{\theta}(a_2|s_2) \times ... \times \pi_{\theta}(a_H|s_H)  \\
    = & \pi_{\theta}(a_1|s_1) \times \pi_{\theta}(a_2|s_1, a_1) \times ... \times \pi_{\theta}(a_H|s_0, a_0, a_1, ..., a_{H-1}) \\
    = & \pi_{\theta}(a_1, a_2, ..., a_H|s_1) \\
    = & \pi_{\theta}(MA \;\;|\;\; MS_1) \\
    = & \pi_{\theta}(MA_1 \;\;|\;\; MS_1) \times \\ 
    & \pi_{\theta}(MA_2 \;\;|\;\; MS_1, MA_1) \times \\
    & ... \\
    & \pi_{\theta}(MA_K \;\;|\;\; MS_1, MA_1, MA_2, ..., , MA_{K-1}) \\
    = & \prod_{T=1}^{K} \pi_{\theta}(MA_T \;\;|\;\; MS_T)
\end{aligned}
\end{equation}
% As mentioned in Equation (\ref{equ:MacroAction2}), $T$ represents the macro timestep.

3. From Equation (\ref{equ:GPG1}) to Equation (\ref{equ:GPG2}), this is similar to the transition from Policy Gradient Theorem to the general form of Policy Gradient Theorem (i.e., from traditinal advantage function to GAE).

\clearpage
\section{The Algorithm Workflow of ARPO}
In this section, we provide a detailed flowchart of the ARPO algorithm in diagram \ref{alg:arpo}.

\begin{algorithm}[H]
\small
\caption{Agentic Reinforce Policy Optimization}
\textbf{Input} initial policy model $\pi_{\theta_{\text{init}}}$; reward models $r_\phi$; task prompts $\mathcal{D}$; 
hyperparameters $\epsilon$, $\alpha$, $\beta$, $\mu$, $\tau$, $M$, $N$, $Z$, $k$
\begin{algorithmic}[1]
\State policy model $\pi_\theta \leftarrow \pi_{\theta_{\text{init}}}$
\For{iteration = 1, \dots, I} \State reference model $\pi_{ref} \leftarrow \pi_{\theta}$
    \For{step = 1, \dots, S}
        % 取数据集
        \State Sample a batch $\mathcal{D}_b$ from $\mathcal{D}$
        % 更新模型
        \State Update the old policy model $\pi_{\theta_{old}} \leftarrow \pi_{\theta}$ 
        % 采样N个输出
        \State Sample $N$ reasoning 1-step paths $\{y_i\}_{i=1}^N \sim \pi_{\theta_{old}} (\cdot \mid q) $ for each question $q \in \mathcal{D}_b$
        % 计算初始熵
        \State Compute initial entropy $H_{i,initial}$ of the first $k$ tokens in each outputs 
        % 定义path集合, 占个位置, $T_{list}$写法有点抽象
        \State let all rollouts $\{rollout_i\} \leftarrow \{y_i\}_{i=1}^N$
        \State tool-call step ${t}$ $ \leftarrow 1$
        % 工具循环
        \While{Any unfinished $y_i$}
            % 执行工具, 拼接results
            \State Parse unfinished $y_i$, execute tools and obtain results $d_i = T(y_i)$
            \State Insert $d$ into rollout $y_i \leftarrow y_i + d_i$
            % 生成额外k个token
            \State Generate $k$ additional tokens based on inserted $y_i$ to compute step-level entropy $H_{i,t}$
            \State Compute normalized change in entropy $\Delta H_{i,t} = Normalize(H_{i,t} - H_{i,inintial})$ for each $y_i$
            % 计算P
            \State Compute partial sampling probability ${P_{i,t}} = \alpha + \beta \cdot \Delta H_{i,t} $
            % 如果P > 
            \If{$P_{i,t} > \tau$}
                \State Branch out $Z$ additional rollouts $\{y_i\}_i^Z$ and add them to $\{rollout_i\}$
            \EndIf
            \If{ $|y_i|=M$} 
                \State Sample $\{y_i\}$ until $y_i$ produce the final answer
            \Else
                \State Sample $\{y_i\}$ to produce new reasoning step
                \State $t \leftarrow t + 1$
            \EndIf
        \EndWhile
        \If{$|y_i| < M$}
            \State Sample $M - |y_i|$ additional independent rollouts and add them to $\{rollout_i\}$
            \State Compute rewards $\{r_i\}_{i=1}^{M}$ for each sampled rollouts $y_i$ by running $r_{\phi}$ 
            \State Compute $\hat{A}_{i,t}$ for the $t$-th token of $o_i$ through group relative advantage estimation.
        \EndIf  
        \For{GRPO iteration = 1, \dots, $\mu$}
            \State Update the policy model $\pi_{\theta}$ by maximizing the GRPO objective (Equation \ref{eq:grpo})
        \EndFor
    \EndFor 
    \State Update $r_\phi$ through continuous training using a replay mechanism. 
\EndFor 
\end{algorithmic}
\textbf{Output} $\pi_\theta$
\label{alg:arpo}
\end{algorithm}

% \clearpage

% \clearpage
\vspace{+10em}
\section{Case Study}

\begin{table*}[!t]
    \centering
    \caption{An example from ARPO on HLE dataset, with special symbols used in think content, search queries, Python codes, returned results and final answer highlighted with \purple{purple box}, \green{green box}, \blue{blue box} and \red{red box}, respectively.}
    \fontsize{9pt}{11pt}\selectfont
    \begin{tabular}{p{0.98\linewidth}}
    \midrule
        \rowcolor{gray!20}\textbf{Example \#1 from HLE Dataset} \\
    \midrule
        \textbf{Question:} \\
       How many non-negative integer solutions are there to the Diophantine equation \(x_1^2 + x_2^2 + x_3^2 + x_4^2 + x_5^2 = 2024\), where \(x_1, x_2, x_3, x_4, x_5 \in \mathbb{Z}_{\ge 0}\)?\\
        \textbf{Labeled Answer:} \\
        29010 \\
    \midrule
        \rowcolor{gray!20}
        \textbf{Model Output by ARPO(14B)}\\
    \midrule
        \purple{<think>} To solve the problem of finding the number of non-negative integer solutions to the equation \(x_1^2 + x_2^2 + x_3^2 + x_4^2 + x_5^2 = 2024\), we can use a brute-force approach with some optimizations. We will iterate over all possible values of \(x_1, x_2, x_3, x_4,\) and \(x_5\) and count the number of times the equation holds true. Here's the Python code to achieve this: 1. We will iterate over all possible values of \(x_1, x_2, x_3, x_4,\) and \(x_5\) from 0 to \(\lfloor \sqrt{2024} \rfloor\), which is 44.   2. For each combination of \(x_1, x_2, x_3,\) and \(x_4\), we will compute \(x_5\) as \(\sqrt{2024 - (x_1^2 + x_2^2 + x_3^2 + x_4^2)}\) and check if it is an integer. Let's write the code to find the solution. \purple{</think>}
        \\\blue{<python>}
        
        \begin{lstlisting}[style=python]
import math
# Initialize the counter for the number of solutions
count = 0
# Iterate over all possible values of x1, x2, x3, x4
for x1 in range(45):
    for x2 in range(45):        
        for x3 in range(45):            
            for x4 in range(45):                
                # Calculate the remaining value for x5^2                
                remaining = 2024 - (x1**2 + x2**2 + x3**2 + x4**2)                
                # Check if the remaining value is a perfect square                
                if remaining >= 0:                    
                    x5 = int(math.isqrt(remaining))                    
                    if x5**2 == remaining:                       
                        count += 1
print(count)
        \end{lstlisting}
        
        \blue{</python>}\\\red{<result>}\\29010\\\red{</result>}
        \\\red{<answer>}\\The number of non-negative integer solutions to the equation \(x_1^2 + x_2^2 + x_3^2 + x_4^2 + x_5^2 = 2024\) is \boxed{29010}.\\\red{</answer>}\\
        
    \bottomrule
    \end{tabular}
    \label{tab:case_gaia3}
\end{table*}

\begin{table*}[!t]
    \centering
    \caption{An example from ARPO on GAIA dataset, with special symbols used in think content, search queries, Python codes, returned results and final answer highlighted with \purple{purple box}, \green{green box}, \blue{blue box} and \red{red box}, respectively.}
    \fontsize{9pt}{11pt}\selectfont
    \begin{tabular}{p{0.98\linewidth}}
    \midrule
        \rowcolor{gray!20}\textbf{Example \#1 from GAIA Dataset} \\
    \midrule
        \textbf{Question:} \\
        I’m researching species that became invasive after people who kept them as pets released them. There’s a certain species of fish that was popularized as a pet by being the main character of the movie Finding Nemo. According to the USGS, where was this fish found as a nonnative species, before the year 2020? I need the answer formatted as the five-digit zip codes of the places the species was found, separated by commas if there is more than one place.\\
        \textbf{Labeled Answer:} \\
        34689 \\
    \midrule
        \rowcolor{gray!20}
        \textbf{Model Output by ARPO(14B)}\\
    \midrule
        \purple{<think>} Okay, let's tackle this question step by step. The user wants to know the five-digit ZIP codes of locations where a particular fish species, which became invasive after being released from pets, was found as a non-native species before 2020. The key clue here is that the fish was the main character in \"Finding Nemo\" and was popularized as a pet because of that movie. First, I need to identify which fish that is. The main character in Finding Nemo is Nemo himself, a clownfish (Amphiprion ocellaris). \textbf{(***More reasoning contentare omitted here***) } Let me start by performing a search to verify the species and its invasive status. First, confirm whether clownfish have indeed become invasive outside their native range due to releases from aquariums. Then check the USGS records for their presence in different regions. So my first search query would be:\purple{</think>}
        \\\green{<search>} Finding Nemo main character invasive fish USGS\green{</search>}\\\red{<result>}\\The USGS Nonindigenous Aquatic Species (NAS) database tracks invasive species occurrences, including non-native fish like the clownfish (*Amphiprion ocellaris*), which is the main character in *Finding Nemo*. While the specific ZIP codes for clownfish sightings before 2020 are not explicitly listed in the provided web pages, the following information is relevant:  1. **Clownfish as Invasive Species**: Clownfish are commonly sold as pets and have been released into non-native environments, such as the Florida Keys (mentioned in Web Page 4).  2. **USGS NAS Database**: The database (Web Pages 1, 2, 6) is the primary resource to query for spatial records of non-native species, including ZIP codes. For example, Web Page 4 notes that ornamental fish releases in Florida have led to invasive species sightings, though specific ZIP codes for clownfish are not provided here.  3. **Florida as a Likely Location**: The Florida Keys National Marine Sanctuary (Web Page 4) has documented non-native fish releases, suggesting ZIP codes in Florida (e.g., 33040 for Key Largo, 33141 for Miami) might be relevant.  To proceed, the NAS database should be queried for *Amphiprion ocellaris* records with dates before 2020 to extract precise ZIP codes.\\\red{</result>} 
        \\\textbf{***Seven search steps omitted here***}
        \\\purple{<think>} Oh! Wait a minute, this changes things. My previous assumption was wrong. The park is actually in Tarpon Springs, FL, which has ZIP code 34689. So the correct ZIP code here is 34689 instead of 33760 or 33761. That's important because using the right ZIP code ensures accuracy. Let me cross-verify this with another source to make sure. Ninth search query: \purple{</think>}
        \\\green{<search>} Tarpon Springs Fred Howard Park ZIP code USGS report\green{</search>}
        \\\red{<result>}\\The non-native *Amphiprion ocellaris* (clownfish) was documented at Fred Howard Park in Tarpon Springs, Florida, in July 2018, according to the USGS Nonindigenous Aquatic Species (NAS) database. The specific ZIP code associated with Fred Howard Park is **34689**, as confirmed by multiple sources including the park’s official address (1700 Sunset Dr, Tarpon Springs, FL 34689) and local ZIP code listings. No other ZIP codes were reported for this species in the USGS records before 2020.  **Five-digit ZIP code:** 34689 \\\red{</result>}\\\purple{<think>}Yes, now I'm confident. The correct ZIP code is 34689. The earlier confusion came from assuming Clearwater, but the park is actually in Tarpon Springs. Therefore, the answer should be 34689. The user asked for five-digit zip codes, and since only one location is found, that's the sole entry. If there were others, we'd list them comma-separated, but according to the latest search, it's just this one.\purple{</think>}\\\red{<answer>}\\So the final answer is the ZIP code 34689.\boxed{34689}\\\red{</answer>}\\
    \bottomrule
    \end{tabular}
    \label{tab:case_gaia1}
\end{table*}

\begin{table*}[!t]
    \centering
    \caption{An example from ARPO on GAIA dataset, with special symbols used in think content, search queries, Python codes, returned results and final answer highlighted with \purple{purple box}, \green{green box}, \blue{blue box} and \red{red box}, respectively.}
    \fontsize{9pt}{11pt}\selectfont
    \begin{tabular}{p{0.98\linewidth}}
    \midrule
        \rowcolor{gray!20}\textbf{Example \#2 from GAIA Dataset} \\
    \midrule
        \textbf{Question:} \\
       In terms of geographical distance between capital cities, which 2 countries are the furthest from each other within the ASEAN bloc according to wikipedia? Answer using a comma separated list, ordering the countries by alphabetical order.\\
        \textbf{Labeled Answer:} \\
        Indonesia, Myanmar \\
    \midrule
        \rowcolor{gray!20}
        \textbf{Model Output by ARPO(14B)}\\
    \midrule
        \purple{<think>} Okay, so I need to figure out which two ASEAN member countries have their capital cities that are the furthest apart geographically. And then present them in alphabetical order, comma-separated. Let me start by recalling what ASEAN stands for—it's the Association of Southeast Asian Nations. The current members include Brunei, Cambodia, Indonesia, Laos, Malaysia, Myanmar (Burma), Philippines, Singapore, Thailand, Vietnam, and maybe others like East Timor or Papua New Guinea? Wait, no, actually I think it's those ten countries. Let me confirm that first. \textbf{(***More reasoning contentare omitted here***)}  First, let me gather the list of capitals and their coordinates. I should search for \"ASEAN member countries and their capitals\" to ensure accuracy. Also, maybe check each country's Wikipedia page for their capital's coordinates. Alternatively, look up a list that includes both names and coordinates. Let me try a search query first.\purple{</think>}
        \\\green{<search>} List of ASEAN countries with capitals and geographic coordinates\green{</search>}\\\red{<result>}\\The searched web pages provide the list of ASEAN member countries along with their capitals but do not include geographic coordinates for the capitals. For example:  - **Brunei**: Capital is **Bandar Seri Begawan**.  - **Cambodia**: Capital is **Phnom Penh**.  - **Indonesia**: Capital is **Jakarta**.  - **Laos**: Capital is **Vientiane**.  - **Malaysia**: Capital is **Kuala Lumpur**.  - **Myanmar**: Capital is **Naypyidaw**.  - **Philippines**: Capital is **Manila**.  - **Singapore**: Capital is **Singapore**.  - **Thailand**: Capital is **Bangkok**.  - **Vietnam**: Capital is **Hanoi**.  To proceed with calculating distances between capitals, geographic coordinates (latitude and longitude) for each capital would need to be obtained from an external source.\\\red{</result>} 
        \\\textbf{***Two search steps omitted here***}
        \\\purple{<think>} So yes, exactly 10 countries. Therefore, the number of pairs is C(10,2)=45. Computing all 45 isn't too bad for a script.\textbf{(***More reasoning contentare omitted here***)} Given all these calculations, I'm confident that the furthest pair is Indonesia and Myanmar. Their capitals are Jakarta and Naypyidaw respectively. Sorting alphabetically, Indonesia comes before Myanmar. Hence the answer should be Indonesia,Myanmar. To determine the two ASEAN member countries with the furthest apart capitals, I first identified the capitals and their geographic coordinates from reliable sources. Using the Haversine formula, I calculated the distances between all pairs of capitals. After performing these calculations, the maximum distance was found between **Jakarta (Indonesia)** and **Naypyidaw (Myanmar)**. When sorted alphabetically, the countries are **Indonesia** and **Myanmar**. \purple{</think>}
        \\\red{<answer>} \\\boxed{Indonesia,Myanmar}
        \\\red{</answer>}\\
    \bottomrule
    \end{tabular}
    \label{tab:case_gaia2}
\end{table*}

\begin{table*}[!t]
    \centering
    \caption{An example from ARPO on HLE dataset, with special symbols used in think content, search queries, Python codes, returned results and final answer highlighted with \purple{purple box}, \green{green box}, \blue{blue box} and \red{red box}, respectively.}
    \fontsize{9pt}{11pt}\selectfont
    \begin{tabular}{p{0.98\linewidth}}
    \midrule
        \rowcolor{gray!20}\textbf{Example \#2 from HLE Dataset} \\
    \midrule
        \textbf{Question:} \\
       How long was the Second Great War in StarCraft Lore (in years, rounded up)\\
        \textbf{Labeled Answer:} \\
        1 \\
    \midrule
        \rowcolor{gray!20}
        \textbf{Model Output by ARPO(14B)}\\
    \midrule
        \purple{<think>}Okay, so I need to figure out how long the Second Great War lasted in StarCraft lore, right? Let me start by recalling what I know about StarCraft's timeline. \textbf{(***More reasoning contentare omitted here***)} So if it starts in 2500 and ends in 2506, that's seven years. But maybe the core conflict is counted until 2505, making it five years. Alternatively, some sources might consider the entire period from the first signs of the Zerg resurgence up to the final battles as spanning six years.\purple{</think>}
        \\\green{<search>}starcraft second great war duration years \green{</search>}\\\red{<result>}\\The Second Great War in *StarCraft* is generally considered to have occurred from **2504 to 2505**, spanning **one year**. This timeline aligns with the events of *StarCraft II: Wings of Liberty* (2504) and *Heart of the Swarm* (2505). However, some sources, such as NamuWiki, specify that the war \"lasted for half a year until April 2505,\" suggesting the **core conflict** may have been shorter (six months), though the broader war period is recorded as one year. The war followed the end of the Brood War (2500) and preceded the *Legacy of the Void* campaign.\\\red{</result>} 
        \\\textbf{***Two search steps omitted here***}
        \\\purple{<think>} Okay, so the official StarCraft Wiki states it's one year. Even if parts of it were shorter, the overall duration is considered a full year. Therefore, rounding up wouldn't change it since it's already a whole number. But just to be thorough, maybe there's a different interpretation. Let me see if any other sources say otherwise.Alternatively, perhaps the Second Great War includes the events leading up to the main games. For example, if the war started in late 2503 and ended in mid-2505, that would be almost two years. But according to the wiki, it's 2504-2505. Given that the user specified \"rounded up,\" even if it were, say, 1.5 years, we'd round to 2. But based on the search results, especially the official wiki, it's one year. So the answer should be 1 year. Wait, but the first search had conflicting info. Let me recheck the exact wording from the wiki.Assuming the official timeline clearly states 2504–2505 as the Second Great War, that's exactly one year. Hence, the duration is one year. Rounded up, it remains one. Therefore, the answer is 1 year. But wait, hold on! Another possibility: the Second Great War might refer to the conflict in the original StarCraft games versus the sequel. No, the Second Great War is definitely part of the StarCraft II storyline. Let me confirm once more.Yes, the Second Great War is indeed the central conflict of StarCraft II, occurring between 2504 and 2505, totaling one year. Some sources might mention the core battle lasting six months, but the entire war period is a year.\purple{</think>}
        \\\red{<answer>} \\ Since the question asks for the duration in years, rounded up, and it's exactly one year, no rounding is needed. \boxed{1}
        \\\red{</answer>}\\
    \bottomrule
    \end{tabular}
    \label{tab:case_gaia4}
\end{table*}

\begin{table*}[!t]
    \centering
    \caption{An example from ARPO on AIME24 dataset, with special symbols used in think content, search queries, Python codes, returned results and final answer highlighted with \purple{purple box}, \green{green box}, \blue{blue box} and \red{red box}, respectively.}
    \fontsize{9pt}{11pt}\selectfont
    \begin{tabular}{p{0.98\linewidth}}
    \midrule
        \rowcolor{gray!20}\textbf{Example \#1 from AIME24 Dataset} \\
    \midrule
        \textbf{Question:} \\
       Let $p$ be the least prime number for which there exists a positive integer $n$ such that $n^{4}+1$ is divisible by $p^{2}$. Find the least positive integer $m$ such that $m^{4}+1$ is divisible by $p^{2}$.\\
        \textbf{Labeled Answer:} \\
        110 \\
    \midrule
        \rowcolor{gray!20}
        \textbf{Model Output by ARPO(8B)}\\
    \midrule
        \purple{<think>} To solve this problem, we need to find the smallest prime \( p \) such that there exists a positive integer \( n \) for which \( n^4 + 1 \) is divisible by \( p^2 \). Then, we need to find the smallest positive integer \( m \) such that \( m^4 + 1 \) is divisible by \( p^2 \). We will start by checking small prime numbers one by one to find the smallest \( p \) that satisfies the condition. We'll use Python to automate this process. First, let's write a function to check if \( n^4 + 1 \) is divisible by \( p^2 \). We'll iterate over the first few values of \( n \) to find such an \( n \) for each prime \( p \). We'll start with the smallest prime number and move upwards until we find the required \( p \). Here's the plan: 1. Iterate over the first few prime numbers. 2. For each prime \( p \), check if there exists an \( n \) such that \( n^4 + 1 \equiv 0 \pmod{p^2} \). 3. Once we find the smallest prime \( p \) that satisfies the condition, we'll find the smallest \( m \) such that \( m^4 + 1 \equiv 0 \pmod{p^2} \). Let's implement this in Python.\purple{</think>}
       \\\blue{<python>}
        
        \begin{lstlisting}[style=python]
from sympy import isprime, primerange
# Function to check if there exists an n such that n^4 + 1 is divisible by p^2
def find_n_for_p(p):
    for n in range(1, p**2 + 1):
        if (n**4 + 1) % (p**2) == 0:
            return n   
        return None
# Iterate over the first few prime numbers to find the smallest p
primes = list(primerange(2, 100))  
# We'll start checking from the first few primes
for p in primes:   
    n = find_n_for_p(p)  
    if n is not None:        
        # Now we need to find the smallest m such that m^4 + 1 is divisible by p^2
        for m in range(1, p**2 + 1):
            if (m**4 + 1) % (p**2) == 0:
                result = m               
                break       
            else:
                continue        
            break
print((p, result))
        \end{lstlisting}
        
        \blue{</python>}
        \\\purple{<think>} The smallest prime \( p \) for which there exists a positive integer \( n \) such that \( n^4 + 1 \) is divisible by \( p^2 \) is \( 17 \). The least positive integer \( m \) such that \( m^4 + 1 \) is divisible by \( 17^2 \) is \( 110 \). Thus, the final answer is: \purple{</think>}
        \\\red{<answer>} \\\boxed{110}
        \\\red{</answer>}\\
    \bottomrule
    \end{tabular}
    \label{tab:case_gaia5}
\end{table*}

\begin{table*}[!t]
    \centering
    \caption{An example from ARPO on HLE dataset, with special symbols used in think content, search queries, Python codes, returned results and final answer highlighted with \purple{purple box}, \green{green box}, \blue{blue box} and \red{red box}, respectively.}
    \fontsize{9pt}{11pt}\selectfont
    \begin{tabular}{p{0.98\linewidth}}
    \midrule
        \rowcolor{gray!20}\textbf{Example \#1 from HotpotQA Dataset} \\
    \midrule
        \textbf{Question:} \\
        Which playwright lived a longer life, Edward Albee or J. M. Barrie?\\
        \textbf{Labeled Answer:} \\
        Edward Franklin Albee III \\
    \midrule
        \rowcolor{gray!20}
        \textbf{Model Output by ARPO(8B)}\\
    \midrule
        \purple{<think>} I need to determine which of two playwrights, Edward Albee and J. M. Barrie, lived a longer life. To do this, I will: 1.  Find the birth date of Edward Albee. 2.  Find the death date of Edward Albee. 3.  Calculate how long Edward Albee lived. 4.  Find the birth date of J. M. Barrie. 5.  Find the death date of J. M. Barrie. 6.  Calculate how long J. M. Barrie lived. 7.  Compare the lifespans. I will use web search for steps 1-5.\purple{</think>}
        \\\green{<search>} Edward Albee birth death dates \green{</search>}
        \\\red{<result>}\\
        Page 1: Sep 16, 2016 · Edward Albee was born Edward Harvey in Washington, D.C. At the age of two weeks, he was adopted by Mr. and Mrs. Reed Albee of Larchmont, New York, and renamed Edward Franklin Albee III. From an early age, … \\
        Page 2: Explore Edward Albee's biography, personal life, family and cause of death. Discover the real story, facts, and details of Edward Albee. \\
        Page 3: Sep 17, 2016 · Edward Albee was a Pulitzer Prize winning American playwright. His popular works include The Zoo Story, and Who’s Afraid of Virginia Wolf? Albee was born Edward Harvey on March 12, 1928 in Washington, D.C. \textbf{(***More search results omitted here***)}
        \\\red{</result>} 
        \\\purple{<think>} From the search results: -   Edward Albee was born on March 12, 1928. -   Edward Albee died on September 16, 2016. Now I need to find J. M. Barrie's birth and death dates.\purple{</think>}
        % \\\green{<search>} J. M. Barrie birth death dates \green{</search>}
        % \\\red{<result>}\\
        % Page 1: Sir James Matthew Barrie, 1st Baronet, (9 May 1860 – 19 June 1937) was a Scottish novelist and playwright, best remembered as the creator of Peter Pan.\\ Page 2: Wikipedia, via WWW, Aug. 20, 2012 (J. M. Barrie; Sir James Matthew Barrie; b. May 9, 1860 in Kirriemuir, Scotland; m. to Mary Ansell, 1894-1909; d. June 19, 1937 in London, England; …\\ 
        % Page 3: Sir James Matthew Barrie, 1st Baronet, (; 9 May 1860 – 19 June 1937) was a Scottish novelist and playwright, best remembered today as the creator of Peter Pan. He was born and educated in … \textbf{(***More search results omitted here***)}
        % \\\red{</result>}
        \\\textbf{***One search step omitted here***}
        \\\purple{<think>} From the search results:-   J. M. Barrie was born on May 9, 1860. - J. M. Barrie died on June 19, 1937. Now I have both sets of dates: \\ Edward Albee: -   Born: March 12, 1928 -   Died: September 16, 2016 \\ J. M. Barrie: -   Born: May 9, 1860 -   Died: June 19, 1937 \\ Next, I will calculate how long each playwright lived. For Edward Albee: 2016 - 1928 = 88 years. For J. M. Barrie: 1937 - 1860 = 77 years. So, Edward Albee lived 88 years and J. M. Barrie lived 77 years. Therefore, Edward Albee lived a longer life.\purple{</think>}
         \\\blue{<python>}
        \begin{lstlisting}[style=python]
# Calculate lifespan for Edward Albee
albee_birth_year = 1928
albee_death_year = 2016
albee_lifespan = albee_death_year - albee_birth_year
print(f"Edward Albee lived for {albee_lifespan} years.")
# Calculate lifespan for J. M. Barrie
barrie_birth_year = 1860
barrie_death_year = 1937
barrie_lifespan = barrie_death_year - barrie_birth_year
print(f"J. M. Barrie lived for {barrie_lifespan} years.")
# Compare lifespans
if albee_lifespan > barrie_lifespan:    
    print("Edward Albee lived a longer life.")
else:    
    print("J. M. Barrie lived a longer life.")
        \end{lstlisting}
        
        \blue{</python>}
        \\\red{<result>} \\ Edward Albee lived for 88 years. J. M. Barrie lived for 77 years. Edward Albee lived a longer life. \\\red{</result>} 
        \\\red{<answer>} \\ \boxed{\text{Edward Albee}}
        \\\red{</answer>}\\
    \bottomrule
    \end{tabular}
    \label{tab:case_gaia6}
\end{table*}

% \begin{small}
% \begin{equation}
% \begin{split}
% J_{\text{GRPO}}(\theta) &= \frac{1}{G} \sum_{i=1}^G \frac{1}{|o_i|} \Bigg[ \sum_{t=1}^{|o_l^i|} \min \Big( r_{i,t}(\theta)^{<l}, \operatorname{clip} \left( r_{i,t}^{<l}(\theta), 1 \pm \epsilon \right) \Big) \hat{A}_{i,t} \\
% & \hspace*{10.5em} + \sum_{t=|o_l^i|}^{|o_i|} \min \Big( r_{i,t}^{>l}(\theta), \operatorname{clip} \left( r_{i,t}^{>l}(\theta), 1 \pm \epsilon \right) \Big) \hat{A}_{i,t} \Bigg] - \beta D_{\text{KL}}(\pi_{\theta} \parallel \pi_{\text{ref}}) \\
% \end{split}
% \label{eq:app grpo before l after l}
% \end{equation}
% \end{small}

\end{document}